%% file: main.tex
\documentclass[10pt,twocolumn,letterpaper]{article}

\usepackage{cvpr}              
\input{preamble}
\usepackage{booktabs}
\usepackage[table]{xcolor}
\definecolor{cvprblue}{rgb}{0.21,0.49,0.74}
\definecolor{mycyan}{RGB}{0,255,255}
\definecolor{mybrown}{RGB}{255, 195, 128}
\usepackage[pagebackref,breaklinks,colorlinks,allcolors=cvprblue]{hyperref}
\definecolor{lightgreen}{RGB}{240,255,255}
\definecolor{grey}{rgb}{0.94,0.94,0.94}
\usepackage{multirow}
\usepackage{tikz}
\usepackage{titletoc}
\usepackage{algorithm}
\usepackage{algorithmic}

\def\paperID{} 
\def\confName{CVPR}
\def\confYear{2026}

\title{\includegraphics[width=0.7cm]{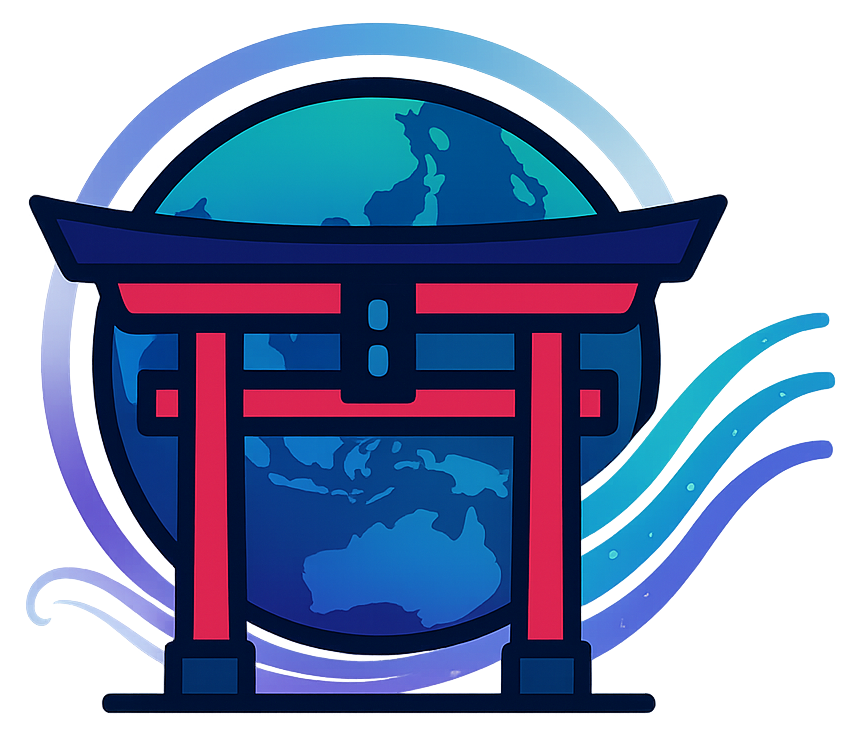} CrossEarth-Gate: Fisher-Guided Adaptive Tuning Engine for Efficient Adaptation of Cross-Domain Remote Sensing Semantic Segmentation}

\author{
    Shilei Cao$^{1,4}$\thanks{Equal contribution.},
    Ziyang Gong$^{3}$\footnotemark[1],
    Hehai Lin$^{7}$, 
    Yang Liu$^{2}$, 
    Jiashun Cheng$^{5}$, \\
    Xiaoxing Hu$^{6}$,
    Haoyuan Liang$^{1,4}$,
    Guowen Li$^{1,4}$, 
    Chengwei Qin$^{7}$,\\
    Hong Cheng$^{2}$,
    Xue Yang$^{3}$,
    Juepeng Zheng$^{1,4}$\thanks{Corresponding author: {\tt\small zhengjp8@mail.sysu.edu.cn}}, 
    Haohuan Fu$^{4,8}$
    \vspace{0.3cm} \\ 
    $^{1}$Sun Yat-sen University,
    $^{2}$The Chinese University of Hong Kong, \\
    $^{3}$Shanghai Jiao Tong University, 
    $^{4}$National Supercomputing Center in Shenzhen, \\
    $^{5}$The Hong Kong University of Science and Technology, 
    $^{6}$Beijing Institute of Technology, \\
    $^{7}$The Hong Kong University of Science and Technology (Guangzhou), 
    $^{8}$Tsinghua University
}

\begin{document}
\maketitle
\input{sec/0_abstract}    
\input{sec/1_introduction}
\input{sec/2_related}

\input{sec/3_method}

\input{sec/4_experiment}

\input{sec/5_conclusion}
\section*{Acknowledgments}
This work was supported in part by the National Natural Science Foundation of China under Grant T2125006 and Grant 42401415; in part by Shenzhen Science and Technology Program under Grant KCXFZ20240903093759004 and Grant KJZD20230923115106012; in part by Guangdong Science \& Technology Program under Grant 2025B0101080001; in part by Tsinghua SIGS KA Cooperation Fund; and in part by the Research Grants Council of the Hong Kong Special Administrative Region, China (No. CUHK 14206625). Yang Liu is supported in part by the Postdoctoral Fellowship Scheme of The Chinese University of Hong Kong.

{
    \small
    \bibliographystyle{ieeenat_fullname}
    \bibliography{main}
}
\input{sec/6_suppl}

\end{document}

%% file: sec/0_abstract.tex
\begin{abstract}
In Remote Sensing (RS), Parameter-Efficient Fine-Tuning (PEFT) has emerged as a key approach to activate the generalizable representation ability of foundation models for downstream tasks. 
However, existing specialized PEFT methods often fail when applied to large-scale Earth observation tasks, as they are unable to fully handle the multifaceted and unpredictable domain gaps (\eg, spatial, semantic, and frequency shifts) inherent in RS data.
To overcome this, we propose CrossEarth-Gate, which introduces two primary contributions.
First, we establish a comprehensive RS module toolbox to address multifaceted domain gaps, comprising spatial, semantic, and frequency modules.
Second, we develop a Fisher-guided adaptive selection mechanism that operates on this toolbox. 
This selection is guided by Fisher Information to quantify each module's importance by measuring its contribution to the task-specific gradient flow. 
It dynamically activates only the most critical modules at the appropriate layers, guiding the gradient flow to maximize adaptation effectiveness and efficiency. 
Comprehensive experiments validate the efficacy and generalizability of our method, where CrossEarth-Gate achieves state-of-the-art performance on 16 out of 18 cross-domain benchmarks for RS semantic segmentation.
\end{abstract}

%% file: sec/1_introduction.tex
\section{Introduction}
With the rapid expansion of Geospatial Foundation Models (GFMs) in both capability and scale~\cite{manas2021seasonal,cong2022satmae,reed2023scale,guo2024skysense}, a central challenge emerges: how to efficiently activate their downstream potential. 
Parameter-Efficient Fine-Tuning (PEFT) \cite{xin2024parameter,zhang2025parameter} methods aim to match or surpass full fine-tuning while updating only a small fraction of parameters, offering an attractive balance between performance and efficiency. 
However, when applied to large-scale and globally distributed Earth observation tasks, existing PEFT methods often suffer substantial performance degradation due to highly variable and unpredictable domain gaps~\cite{hu2025earth,wang2021loveda,liu2023large,liang2025low}.



\begin{figure*}[tbp]
\centering
\includegraphics[width=\textwidth]{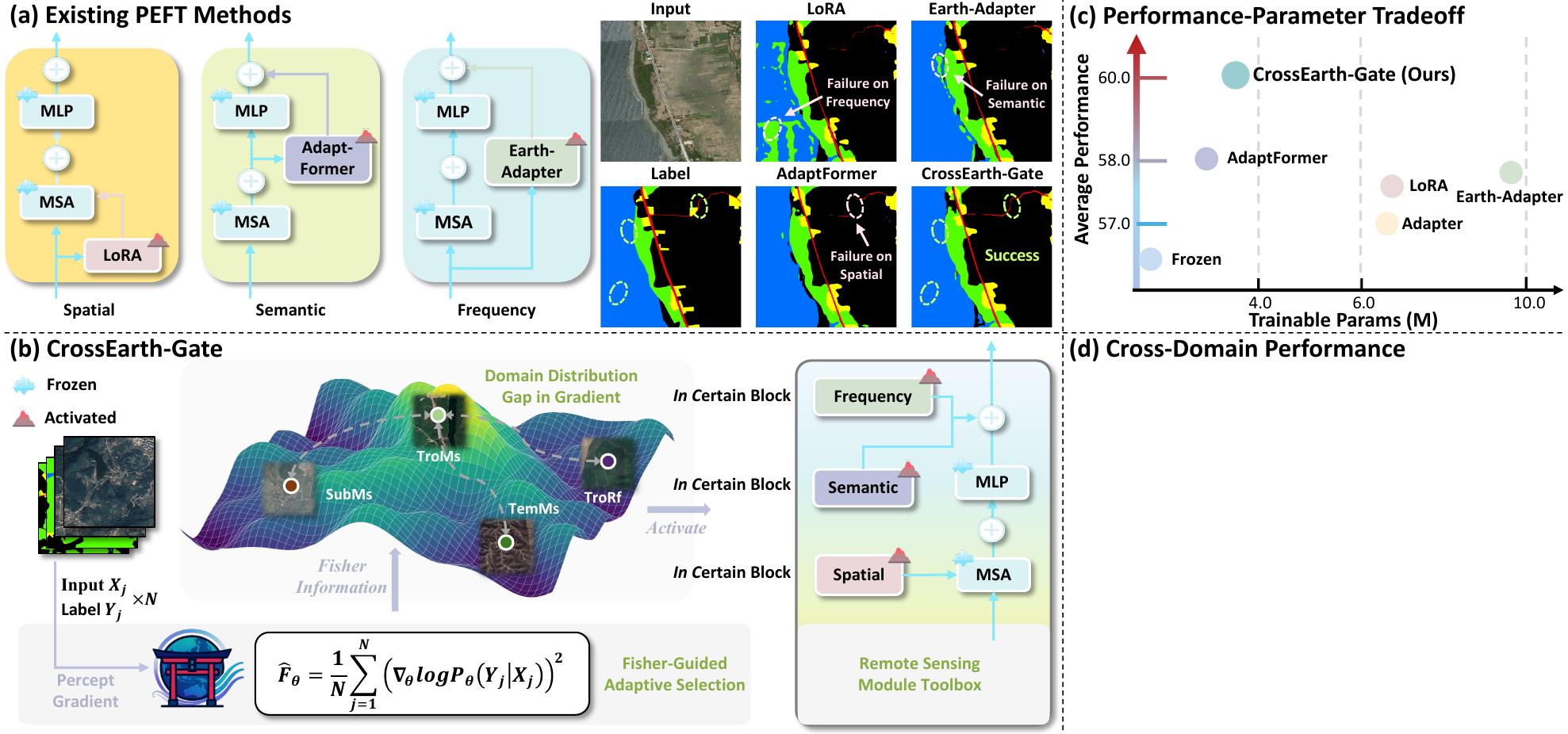} 

\begin{tikzpicture}[overlay,remember picture]
    \node[anchor=south east, xshift=-1.95cm, yshift=16.79cm] at (current page.south east) {
        \includegraphics[width=0.32\textwidth]{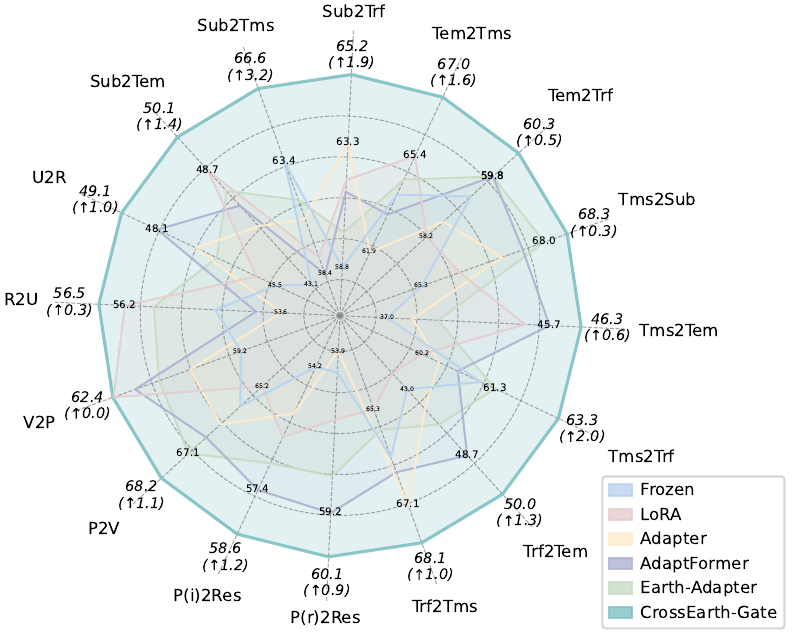} 
    };
\end{tikzpicture}
\vspace{-9pt}
\caption{
Overview of the CrossEarth-Gate and its comparative advantages. 
(a) Existing PEFTs typically focus on one specific functional pathway (\eg, LoRA for spatial, Adaptformer for semantic, Earth-Adapter for frequency).
The qualitative example of generalization across different climate zones shows each baseline failing on challenges outside its specialty, while our method succeeds.
(b) Our proposed CrossEarth-Gate establishes a toolbox combining all three module types. 
Then, we utilize Fisher Information to guide the gradient flow to activate only the most critical modules at the most relevant blocks for a specific domain.
(c) CrossEarth-Gate results in a superior Performance-Parameter Tradeoff. 
(d) CrossEarth-Gate achieves state-of-the-art results across 16 challenging DG and DA benchmarks.
}
\vspace{-1pt}
\label{fig:motivation}
\end{figure*}

Domain gaps in Remote Sensing (RS) are multifaceted, arising from diverse sources like wavelength ranges, geographical landscapes, and climatic zones.
These gaps manifest as a complex interplay of adaptation challenges: 
(1) \textbf{Spatial shifts}, representing the change in object structure and scale that require geometric integrity; 
(2) \textbf{Semantic shifts}, denoting the differences in class appearance and concepts; 
and (3) \textbf{Frequency shifts}, involving high-frequency spectral artifacts or textural noise from different features.
However, as shown in Fig.~\ref{fig:motivation} (a), existing PEFTs, whether general-purpose or RS-specific, typically specialize in one functional pathway.
For example, LoRA \cite{hu2022lora} modifies the Multi-Head Self-Attention (MSA) layers to enhance spatial dependency modeling.
AdaptFormer \cite{chen2022adaptformer} adjusts the Multi-Layer Perception (MLP) layers to refine high-level semantic features.
The Earth-Adapter \cite{hu2025earth} focuses on mitigating high-amplitude artifacts in the frequency domain.


While each method has its merits, this single-pathway design is inherently limited: it captures only one facet of RS domain shifts while leaving others unaddressed. 
As shown by the segmentation maps in Fig.~\ref{fig:motivation}(a), LoRA misclassifies large areas of water with high-frequency waves as forest, revealing its inability to process unfamiliar frequency artifacts.
AdaptFormer fails to maintain spatial continuity of the road, shattering its structure and demonstrating a critical failure in spatial awareness.
Earth-Adapter falters on semantic ambiguities and misidentifies coastal waves as forest, committing a semantic error.
These dilemmas motivate us to pursue a more dynamic mechanism capable of jointly addressing spatial, semantic, and frequency variations.

To address this, we propose \textbf{CrossEarth-Gate}, a Fisher-\textbf{G}uided \textbf{a}daptive \textbf{t}uning \textbf{e}ngine for \textbf{Cross}-domain PEFT in \textbf{Earth} observation.
Our methods provide a dynamic and comprehensive solution to multifaceted domain gaps, which comprises two components.
First, CrossEarth-Gate establishes a structured \textbf{RS module toolbox}, designed to inject trainable parameters into a different functional pathway of models, including spatial, semantic, and frequency modules using different PEFT methods.
This provides a unified framework with the full capacity to address complex domain shifts. 
Another novel and critical design is the \textbf{Fisher-guided adaptive selection} that operates on this toolbox to dynamically choose the most critical modules at the appropriate layers. 
Specifically, Fisher Information guides the selection \cite{fisher1922mathematical}, which serves as a metric to evaluate the importance of each module in terms of its contribution to the current task-specific gradient flow.
This mechanism periodically determines where to ``cast its hook'', directing the gradient flow of the model’s parameters in a way that maximizes adaptation efficiency and effectiveness for domain-specific shifts.
This process facilitates a multi-stage learning process, allowing for a robust and nuanced adaptation.

In addition, we rigorously evaluate CrossEarth-Gate across 18 Domain Generalization (DG) and Domain Adaptation (DA) benchmarks for RS semantic segmentation. 
It achieves State-Of-The-Art (SOTA) results on 16 of them, outperforming PEFT baselines by up to 3.2\% mIoU while offering a better performance-efficiency tradeoff (See Fig. \ref{fig:motivation} (c) and (d)).
The main contributions of this work are:
\begin{itemize}
    

    \item We address the challenges posed by extreme heterogeneity and multifaceted domain shifts inherent in RS adaptation tasks with a unified, comprehensive framework.

    \item We propose CrossEarth-Gate, which introduces a structured RS module toolbox to comprehensively tackle the multifaceted RS domain shifts and a Fisher-guided adaptive selection mechanism to dynamically guide the gradient flow to maximize efficiency and effectiveness.
    
    \item CrossEarth-Gate outperforms existing specialized cross-domain methods, FMs, and PEFE approaches across 16 cross-domain benchmarks for RS semantic segmentation, validating its strong generalizability and effectiveness.
\end{itemize}

%% file: sec/2_related.tex
\section{Related Works}

\subsection{Geospatial Foundation Models}
The concept of Foundation Model (FM) originates from the NLP \cite{wiggins2022opportunities} domain, defined as ``the base models trained on large-scale unlabeled data that can be adapted for a variety of downstream tasks'' \cite{vaswani2017attention,devlin2019bert,raffel2020exploring,brown2020language,radford2021learning,chowdhery2023palm,touvron2023llama,wei2023cat}.
Driven by the success of self-supervised learning (SSL) techniques such as masked patch reconstruction, Vision Foundation Models (VFMs) have yielded significant achievements in Computer Vision (CV), serving as powerful backbones for general visual feature representation \cite{he2022masked, oquab2023dinov2, chen2020simple, he2020momentum, kirillov2023segment, dosovitskiy2020image}.

Adapting existing VFM architectures and techniques to Earth observation data, typically represented as images, has given rise to Geospatial Foundation Models (GFMs) \cite{manas2021seasonal,tang2023cross,guo2024skysense,cong2022satmae,reed2023scale,wang2025hypersigma,wang2024feature,scheibenreif2023masked,li2025saratr,gong2024crossearth,liu2025cirt,li2026tianquanss,zhang20262}.
For example, SatMAE \cite{cong2022satmae} introduces temporal and multi-spectral positional encodings for satellite imagery within the mask image modeling pre-training.
To tackle domain shift, CrossEarth \cite{gong2024crossearth} utilizes Earth-style injection and multi-task training for domain generalization.
However, as the representation capabilities and scale of these FMs increase, how to efficiently and effectively leverage their potential for diverse RS downstream tasks remains a key problem.

\subsection{Cross-Domain PEFT in RS}
PEFT has emerged as a promising solution to efficiently fine-tune FMs while using a minimum number of trainable parameters \cite{houlsby2019parameter,hu2022lora,jia2022visual,han2024parameter,zhang2025parameter}. 
These methods can be broadly classified into four categories: Selective, Additive, Prompt, and Reparameterization approaches \cite{zhang2025parameter}.
Selective PEFT focuses on only optimizing a specific subset of the model's parameters \cite{zhao2023tuning,zaken2021bitfit,sung2021training,ansell2021composable,zhao2024sct,xu2021raise,cao2026taskadaptive}.
For instance, \cite{sung2021training} selects sparse parameters based on Fisher information as a measure of parameter importance. 
Additive PEFT introduces extra trainable parameters into the frozen backbone \cite{houlsby2019parameter,pfeiffer2020adapterfusion,chen2022adaptformer,pfeiffer2020adapterfusion,yin20255}.
Prompt PEFT incorporates learnable prompts into the input or the attention layers to adapt to a specific task.
Reparameterization PEFT reformulates or decomposes existing model parameters so that only a subset requires adjustment during fine-tuning \cite{hu2022lora,liu2021enabling}.
Particularly, LoRA \cite{hu2022lora} decomposes the updated weight into two low-rank matrices, inspiring numerous subsequent methods \cite{dettmers2023qlora,zhang2023lora,liu2024dora,hayou2024lora+,zhong2024convolution}.
Moreover, existing cross-domain studies begin to explore PEFT for Domain Adaptation (DA) \cite{gao2022visual,sun2023vpa,gong2024coda,yang2024exploring} and Domain Generalization (DG) \cite{hu2024learn,wei2024stronger}.

Recently, PEFT techniques have also gained increasing attention within the RS community, which commonly leverages VFMs or GFMs for various downstream tasks \cite{dong2024upetu,hu2024tea,hu2024airs,scheibenreif2024parameter,zhang2025spectralx,thoreau2025parameter,yin2025remote}. 
For instance, SLR \cite{scheibenreif2024parameter} and DEFLECT \cite{zhang2025parameter} utilize low-rank matrices to adapt to different modalities of RS data, while Earth-Adapter \cite{hu2025earth} is designed to mitigate the domain shift resulting from frequency-domain artifacts.
However, these PEFT methods typically focus on one aspect of the problem, such as spatial, semantic, and frequency adaptation, failing to address the complex, multifaceted nature of RS domain shifts. 
A method specialized for one aspect often fails when confronted with challenges from another.
Consequently, a single static method fails to address these intertwined challenges comprehensively.
To address this, we establish a structured toolbox combining three module types, targeting different functional pathways of the model.
We then propose a Fisher-guided adaptive selection mechanism that dynamically identifies the most relevant modules, enabling a more comprehensive cross-domain adaptation.

%% file: sec/3_method.tex
\section{Preliminary}
\paragraph{Optimization Objective.}
Let us consider a cross-domain adaptation task with a training dataset $D = \{(\mathbf{X}, \mathbf{Y})\}$, where the $\mathbf{X} \in \mathbb{R}^{h \times w \times c}$ denotes the input RS images, comprising $h \times w$ pixels with $c$ bands, and $\mathbf{Y}$ represents corresponding label. 
For Domain Generalization (DG) tasks, $D$ represents the labeled source domain dataset, while Domain Adaptation (DA) additionally includes the unlabeled target domain input and its pseudo-labels.
Given a pretrained FM with parameters $\alpha$, PEFT-related modules with parameters $\zeta$ ($|\zeta|\ll|\alpha| $), and the decoder head $\mathcal{H}_\phi$ parameterized by $\phi$, the objective of cross-domain PEFT is to enhance the model performance in the target domain by optimizing:
\begin{equation}
\arg\min_{\zeta,\phi}\mathbb{E}_{(\mathbf{X},\mathbf{Y}) \in D} \mathcal{L}(\mathcal{H}_\phi(\mathcal{B}_{\alpha,\zeta}(\mathbf{X})),\mathbf{Y}), 
\end{equation}
where $\mathcal{B}_{\alpha,\zeta}$ signifies the backbone incorporating the PEFT modules and $\mathcal{L}$ is the loss function for the adaptation task. 

\paragraph{Transformer Block.} 
This work mainly focuses on Transformer-based \cite{vaswani2017attention} FMs, such as ViT \cite{dosovitskiy2020image}.
The input $\mathbf{X}$ is first divided into $l$ patches and then encoded into $d$-dimensional embedding through a patch embedding layer, yielding the initial tokens $\mathbf{T}_1 \in \mathbb{R}^{l \times d}$.
Subsequently, $\mathbf{T}_1$ is processed by a sequence of $I$ transformer blocks in the FM. 
We denote the input features of $i$-th block as $\mathbf{T}_{i} \in \mathbb{R}^{l \times d}$.
Each block is typically composed of a Multi-head Self-Attention (MSA) module and a feed-forward Multi-Layer Perceptron (MLP) network, with residual connections and Layer Normalization (LN) applied after each module.
The computation for the $i$-th block is defined as follows:
\begin{align}
\mathbf{T}^{\text{attn}}_{i} &= \text{MSA}(\mathbf{T}_{i}) + \mathbf{T}_{i}, \quad \mathbf{T}^{\text{attn}}_{i} \in \mathbb{R}^{l \times d}, 
\label{msa}
\\
\mathbf{T}_{i+1} &= \text{MLP}(\mathbf{T}^{\text{attn}}_{i}) + \mathbf{T}^{\text{attn}}_{i}, \quad \mathbf{T}_{i+1} \in \mathbb{R}^{l \times d},
\label{mlp}
\end{align}
where we omit normalization for simplicity and $\mathbf{T}_{i+1}$ is the output of $i$-th block, which is passed as input to next block. 

\section{Methods}
The Fig. \ref{fig:motivation} (b) presents an overview of the CrossEarth-Gate, which consists of a structured RS module toolbox and Fisher-guided adaptive selection mechanism.

\subsection{Remote Sensing Module Toolbox}
First, to equip the model with the comprehensive capabilities needed to tackle intertwined domain gaps, CrossEarth-Gate first establishes a structured and comprehensive \textbf{RS module toolbox}. 
This toolbox integrates spatial, semantic, and frequency modules, each engineered into distinct functional pathways to provide specific aspects of the feature hierarchy for specific domain challenges. 
Instead of statically choosing one method for certain layers, we initially insert all these modules at every layer of the pre-trained backbone.
\paragraph{Spatial Module.}
The MSA sub-layer (Eq. \eqref{msa}) explicitly models the relationships between tokens, allowing the model to build contextual understanding and capture spatial dependencies at various scales \cite{luo2024understanding,hong2023context}. 
In the RS application scenarios, adapting to change in object scale, spatial arrangement, or geographic layout necessitates a targeted tuning of this reasoning mechanism.
Therefore, we employ the Spatial Module using Low-Rank Adaptation (LoRA) \cite{hu2022lora}. 
LoRA is predicated on the hypothesis that the weight update $\Delta \mathbf{W}$ for a pre-trained weight matrix $\mathbf{W}_0 \in \mathbb{R}^{d \times d}$ during adaptation possesses a low ``intrinsic rank''. 
This update can be parameterized by the product of two low-rank matrices, $\mathbf{A} \in \mathbb{R}^{d \times r}$ and $\mathbf{B} \in \mathbb{R}^{r \times d}$, where $r \ll d$ is the rank.
The merged weight $\mathbf{W}$ is written as: 
\begin{equation}
    \mathbf{W} = \mathbf{W}_0 + \Delta \mathbf{W} = \mathbf{W}_0 + \mathbf{BA}, \quad \mathbf{W} \in \mathbb{R}^{d \times d}.
\end{equation}
We strategically inject these trainable, low-rank matrices into the query ($\mathbf{W}_Q$) and value ($\mathbf{W}_V$) linear projection weights of each MSA module (Eq. \eqref{msa}). 
This targeted intervention directly modulates the self-attention mechanism, influencing how the model weighs and aggregates spatial information.
It allows the model to adapt its understanding of spatial context and object scales without altering the parameters in the original, frozen $\mathbf{W}_Q$ and $\mathbf{W}_V$ weights. 

\paragraph{Semantic Module.}
The MLP sub-layer (Eq. \eqref{mlp}) transforms each token's representation independently, which is widely understood to be a key locus of factual and semantic knowledge within the model \cite{geva2020transformer,meng2022locating,yao2024knowledge}. 
Moreover, generalizing concepts across different domains (e.g., ``building'' in rural and urban landscapes) logically requires modifying this stored knowledge.
To adapt the model's high-level semantic knowledge, we introduce a Semantic Module based on the Adapter architecture \cite{houlsby2019parameter,chen2022adaptformer}, placed in parallel with each MLP sub-layer.
The adapter in $i$-th block comprises a down-projection layer with weight $\mathbf{W}_i^{down} \in \mathbb{R} ^ {d \times \hat{d}}$, a GELU activation, and an up-projection layer with weight $\mathbf{W}_i^{up} \in \mathbb{R} ^ {\hat{d} \times d}$, satisfying $\hat{d} \ll d$.
The output of this module is added to the original MLP's output via a residual connection. 
Given the modified output $\hat{\mathbf{T}}^{\text{attn}}_{i}$ of the MSA integrating LoRA, the modified output of block  $\hat{\mathbf{T}}_{i+1}$ becomes:
\begin{gather}
\hat{\mathbf{T}}_{i+1} = \text{MLP}(\hat{\mathbf{T}}^{\text{attn}}_{i}) + \text{Adapter}_i(\hat{\mathbf{T}}^{\text{attn}}_{i}) + \hat{\mathbf{T}}^{\text{attn}}_{i}, \\
\text{Adapter}_i(\hat{\mathbf{T}}^{\text{attn}}_{i})= \text{GELU}(\hat{\mathbf{T}}^{\text{attn}}_{i} \cdot\mathbf{W}_i^{down}) \cdot \mathbf{W}_i^{up}.
\end{gather}
This parallel structure allows the module to refine or adjust the semantic transformation performed by the frozen MLP without disrupting the original pre-trained knowledge flow.

\paragraph{Frequency Module.}
RS imagery is uniquely challenged by high-amplitude artifact influences, which are almost situated everywhere in the RS image.
To address this, we introduce Frequency Modules based on the Earth-Adapter \cite{hu2025earth}, which operates in the frequency domain. 
The module first utilizes the Fourier Transform to decompose the input features into low-frequency (structural) and high-frequency (detail/texture) components.
These disentangled components are processed by distinct, lightweight adapter experts.
A mixture-of-adapters router then learns to selectively process and recombine these frequency components, effectively mitigating artifact disturbances while preserving essential features.
Given the modified output of the $i$-th block $\hat{\mathbf{T}}_{i+1}$, the input to next block $\tilde{\mathbf{T}}_{i+1}$ becomes:
\begin{equation}
    \tilde{\mathbf{T}}_{i+1} =  \hat{\mathbf{T}}_{i+1} + \text{Earth-Adapter}_i(\hat{\mathbf{T}}_{i+1}).
\end{equation}

\subsection{Fisher-Guided Adaptive Selection}

Given the toolbox, the central challenge becomes determining which modules to activate for a specific task. 
Tuning all modules simultaneously is fundamentally inefficient, yet a static or heuristic selection would fail to efficiently leverage the full potential of the toolbox for diverse domain shifts, reverting to a sub-optimal strategy. 
Therefore, CrossEarth-Gate employs a principled, data-driven selection mechanism as the dynamic guide.
We conceptualize the adaptation as a gradient flow: as the model fine-tunes, task-specific gradients flow through the network, and our different module types offer distinct pathways for this flow. 
Our goal is to periodically analyze this flow and   ``gate'' it, directing it only to the modules that offer the highest impact for adaptation.

To quantify this impact in a theoretically-grounded way, 
we turn to the Fisher Information Matrix (FIM) \citep{fisher1922mathematical}.
A parameter’s significance can be determined by evaluating the extent to which altering it influences the model's output.
Given a model parameterized by $\theta \in \mathbb{R}^{|\theta|}$, we denote the output distribution of the model as $P_{\theta}(\mathbf{Y}|\mathbf{X})$ for input $\mathbf{X}$.
Subsequently, we can assess how much a small parameter perturbation $\delta \in \mathbb{R}^{|\theta|}$ in the parameter changes the distribution using the Kullback-Leibler divergence \cite{kullback1951information} $\text{D}_{KL}(P_{\theta}(\mathbf{Y}|\mathbf{X}) \parallel P_{\theta+\delta}(\mathbf{Y}|\mathbf{X}))$.
As shown in \cite{abbass2022review,pascanu2013revisiting}, when $\delta \rightarrow 0$, the following second-order approximation holds: 
\begin{equation}
\mathbb{E}_{\mathbf{X}}\left [ \text{D}_{KL}(P_{\theta}(\mathbf{Y}|\mathbf{X}) \parallel P_{\theta+\delta}(\mathbf{Y}|\mathbf{X})) \right] = \delta^{T} F_{\theta} \delta + O(\delta^{3}),
\end{equation}
where $F_{\theta} \in \mathbb{R}^{|\theta|\times|\theta|}$ is the FIM \citep{fisher1922mathematical}, defined as:
\begin{equation}
F_{\theta} = \mathbb{E}_{\mathbf{X}} \left [ \mathbb{E}_{\mathbf{Y} \sim P_{\theta}(\mathbf{Y}|\mathbf{X})} \nabla_{\theta} \text{log}P_{\theta}(\mathbf{Y}|\mathbf{X}) \nabla_{\theta} \text{log}P_{\theta}(\mathbf{Y}|\mathbf{X})^{T} \right].
\label{eqF}
\end{equation}
Apparently, the FIM links $\delta$ to the resultant changes in the model’s output distribution.
However, the $|\theta| \times |\theta|$ size of $F_{\theta}$ renders its exact computation infeasible. 
Consequently, we adopt the empirical diagonal approximation of the FIM \cite{kirkpatrick2017overcoming,sung2021training}, which is equivalently a vector in $\mathbb{R}^{|\theta|}$.
When sampling $N$ data pairs $(\mathbf{X}_j, \mathbf{Y}_j)$ from the task dataset $D$, the diagonal FIM for $\theta$ can be empirically estimated as:
\begin{equation}
\hat{F_{\theta}} = \frac{1}{N} \sum_{j=1}^{N} (\nabla_{\theta} \text{log}P_{\theta}(\mathbf{Y}_j|\mathbf{X}_j))^2,
\label{eqFhat_practical}
\end{equation}
where $\text{log}P_{\theta}(\mathbf{Y}_j|\mathbf{X}_j)$ is realized as the negative task loss.
This practical approximation relates the $\hat{F_{\theta}}$ to the square gradient of the parameter. 
We can conceptualize the adaptation as a flow, where the task-specific loss directs gradients back to the parameters. 
The gradient represents the direction and magnitude of this flow. 
A large $\hat{F_{\theta}}$ value signifies that this parameter is a ``high-flow'' channel, \ie, a pathway where gradients are large and thus highly impactful on the model's output. 
This is precisely where we should ``cast our hook'' to achieve the maximum adaptation gain.

The FIM approximation in Eq. \eqref{eqFhat_practical} provides an importance score for an individual parameter.
To select entire modules, CrossEarth-Gate employs a dynamic gating mechanism that periodically re-evaluates module importance throughout the fine-tuning process.
By using the empirical FIM as our ``flow meter'', we are able to dynamically ``gate'' the gradient flow, directing it only to the modules that offer the highest impact.
Specifically, at every $N$ training iterations, the framework will temporarily activate all modules in the toolbox. 
The Fisher information is then computed by summing over a small batch of $M$ data samples.
To derive a module-level importance score $S_i^z$ for the $z$-th type of module parameterized by $\zeta_i^z$ in the $i$-th block, we aggregate the scores of all its constituent parameters:
$\hat{S_i^z}=\sum_{\zeta_i^z} \hat{F}_{\zeta_i^z}$. 
To make the scores of different types of modules comparable, we compute the category-normalized relative importance scores:
$
S_i^z = \frac{ \hat{S_i^z}}{ \sum_i^I \hat{S_i^z}}
$, where $I$ represents the number of blocks.
This ensures that the selection process is balanced, making all module types comparable and promoting a diverse, task-specific configuration.
Subsequently, only Top-k modules with the highest scores are activated, and the gradient flow is gated to only these pathways for the next $N$ training iterations.
CrossEarth-Gate repeats this process to dynamically adapt its tuning strategy to the most critical components for the task at hand.
The hyperparameter configurations are listed in the Appendix.

%% file: sec/4_experiment.tex
\input{sec/tab/tab1}

\section{Experiments}

\subsection{Settings}
We conduct a comprehensive evaluation of CrossEarth-Gate on DG and DA benchmarks for RS semantic segmentation tasks. 
Additional dataset details, implementation specifics, and baseline configurations are provided in the Appendix.
\paragraph{Datasets.}
Our cross-domain analysis leverages a suite of 
diverse datasets.
The CASID dataset specifically addresses domain shifts across four climate zones: Subtropical Monsoon (Sub), Temperate Monsoon (Tem), Tropical Monsoon (Tms), and Tropical Rainforest (Trf).
The ISPRS Potsdam and Vaihingen contain aerial images collected over the cities of Potsdam and Vaihingen, with IR-R-G, R-G-B, and R-G-B-IR channels.
RescueNet consists of aerial imagery focused on detecting buildings impacted by disasters to facilitate rescue operations.
LoveDA presents a cross-scene dataset with RS images from both urban and rural areas.
For DG experiments, we follow the protocol from \cite{hu2025earth}.
We employ CASID to establish 12 out-of-domain generalization scenarios. 
We also evaluate on RescueNet with source domain as the RGB (P(r)2Res) and IR–R–G (P(I)2Res) channels of Potsdam.
For DA experiments, we form four benchmarks: Potsdam to Vaihingen (P2V), Vaihingen to Potsdam (V2P), Rural to Urban (R2U), and Urban to Rural (U2R).
These datasets provide a spanning multifaceted RS domain shifts, including unseen geographical regions, novel spectral bands, and diverse climatic conditions.

\paragraph{Implementation Details.}
All experiments are implemented based on the MMSegmentation \cite{mmseg2020} framework.
We mainly employ Dinov2 \cite{oquab2023dinov2} as the feature extraction backbone, paired with Mask2Former \cite{cheng2022masked} as the segmentation head.
For DA experiments, we utilize the DACS \cite{tranheden2021dacs} self-training framework, whereas for DG experiments, we train the model via end-to-end supervised learning.
The models are finetuned using AdamW \cite{loshchilov2017decoupled} optimizer with a base learning rate of 1e-5 for the decoder and PEFT modules.

\paragraph{Baselines.}
We compare CrossEarth-Gate against existing PEFT methods, including VPT \cite{jia2022visual}, SLR \cite{scheibenreif2024parameter}, Rein \cite{wei2024stronger}, LoRA \cite{hu2022lora}, Adapter \cite{houlsby2019parameter}, Adaptformer \cite{chen2022adaptformer}, and Earth-Adapter \cite{hu2025earth}.
We also select two conventional fine-tuning approaches: ``Frozen'', where all backbone parameters are fixed, and ``Full-Tuning'', where we fine-tune the entire model.
Additionally, for the DG experiments, we include results from two cross-domain specialized models and the Full-Tuning of different GFMs, based on \cite{gong2024crossearth}.

\subsection{Cross-Domain Performance Comparison}

\paragraph{Generalization Across Climate Zones.}
As presented in Tab. \ref{exp:casid}, Full-Tuning undergoes a catastrophic performance decline compared to the Frozen backbone, confirming that unconstrained fine-tuning severely overfits to the source domain. 
Specialized, static PEFTs present inconsistent performance, showing varying efficacy depending on the domain.
For instance, the Earth-Adapter achieves the second-best average result (64.5\%) when trained on the Tem domain, whereas the Adaptformer performs better (61.8\%) when trained on the Trf domain.
This variance suggests that different climate-zone transfers necessitate a different emphasis on spatial, semantic, or frequency adjustments, exposing the inherent limitations of the specialized PEFT strategy.
In contrast, CrossEarth-Gate demonstrates consistent performance, achieving SOTA results in 10 of the 12 scenarios and securing the highest average mIoU across all four source domains, all while maintaining high parameter efficiency.
This superior result highlights the critical advantage of our structured RS module toolbox, which comprehensively tackles the complex RS domain shifts.
While our method is dominant, in two isolated cases (Tem2Sub and Trf2Sub), static methods like Earth-Adapter or Adaptformer achieve marginally better results. 
This suggests that for these specific shifts, their fixed module placement coincidentally aligns well with the required feature adaptation. 
However, these static, specialized methods fail to generalize to other domains.
The comprehensive results underscore the robustness and superior generalization capability of the CrossEarth-Gate framework across different climate zones.
\input{sec/tab/tab2}
\input{sec/tab/tab3}


\paragraph{Generalization to Disaster Scenarios.}
Tab. \ref{exp:rescue} shows the results of the generalization task from a standard aerial dataset (Potsdam) to a post-disaster scenario (RescueNet), involving simultaneous shifts in geography, spectral bands, and target object appearance. 
CrossEarth-Gate again achieves the highest mIoU in both experiments, with an mIoU of 60.1\% on P(r)2Res and 58.6\% on P(i)2Res. 
Our method demonstrates strong generalization on ambiguous, large-scale classes like Impervious surfaces and Clutter, which are prevalent in post-disaster environments.
However, for the Building class, the Frozen backbone outperforms our method in the P(r)2Res case. 
This is a noteworthy finding, suggesting that the DINOv2's pre-trained representations for highly structured objects are already extremely robust and generalizable. 
In this context, any adaptation may marginally disrupt these near-optimal features for the Building class. 
Despite these class-specific variances, the comprehensive SOTA mIoU scores validate the superior and more balanced generalization capability of CrossEarth-Gate in the complex cross-domain RS scenarios.


\paragraph{Adaptation to Unlabeled Target Domains.}
The DA results presented in Tab.~\ref{exp:da}, where the model has access to unlabeled target data, further validate our approach. 
Full-Tuning fails again, likely due to instability from noisy target-domain pseudo-labels.
In contrast, PEFT methods, including LoRA, Adaptformer, and Earth-Adapter, all outperform the strong Frozen baseline, demonstrating their benefits in the DA context.
However, CrossEarth-Gate achieves the best performance across four adaptation scenarios with the highest average mIoU of 59.1\%. 
This consistent success highlights the efficacy of our Fisher-guided selection in the DA context. 
The gradient flow, informed by pseudo-labels from a specific target domain, allows our model to precisely identify and activate only the most critical modules needed for that particular adaptation.
While our method tied with the LoRA on the V2P task, this does not indicate a limitation. Rather, it suggests that this specific V2P adaptation may be predominantly a spatial-scale challenge, which LoRA's static architecture coincidentally addresses well. 
However, CrossEarth-Gate's ability to win or tie across all diverse UDA scenarios validates its more robust and principled approach to dynamic adaptation.

\begin{figure*}[t]
    \centering
    \includegraphics[width=\linewidth]{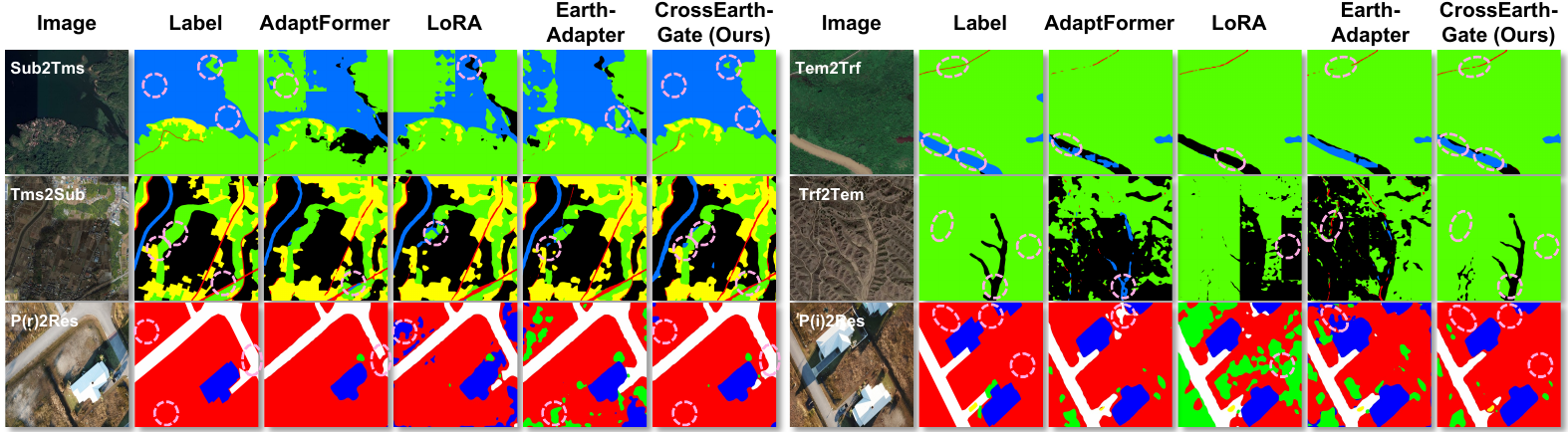}
    \vspace{-12pt}
    \caption{
    Visualizations of predicted segmentation maps of PEFT methods. 
    In the CASID \cite{liu2023large} dataset, {\color{red} red} is the road class, {\color{yellow} yellow} is the building class, {\color{blue} blue} is the water class, {\color{green} green} is the forest class, and black is the background class.
    In the RescueNet \cite{rahnemoonfar2023rescuenet} dataset, white is the impervious surface class, {\color{red} red} is the clutter class, {\color{blue} blue} is the building class, {\color{green} green} is the vegetation class, and {\color{yellow} yellow} is the car class.
    }
    \vspace{-4pt}
    \label{fig:vis}
\end{figure*}

\subsection{Ablation Studies}
\paragraph{Backbone Generalizability.}
As detailed in Tab.~\ref{exp:ablation}, we evaluate CrossEarth-Gate across five pre-trained VFMs and GFMs on the CASID benchmark, including DINOv2 series \cite{oquab2023dinov2}, SAM \cite{kirillov2023segment}, SatMAE \cite{cong2022satmae}, and Scale-MAE \cite{reed2023scale}.
Specifically, Full-Tuning is an unstable and generally poor strategy for DG. 
It performs significantly worse than the Frozen baseline on the DINOv2 series, a clear case of overfitting to the source domain, while incurring an enormous parameter cost. 
In contrast, CrossEarth-Gate consistently and significantly outperforms both the Frozen and Full-Tuning baselines across all five architectures. 
It achieves the highest average mIoU in every case, with improvements from 2.1 \% to 7.6 \% mIoU, while tuning a minimal number of parameters. 
This robust performance strongly validates the remarkable versatility and generalizability of the CrossEarth-Gate.

\input{sec/tab/tab4}

\begin{figure}[t]
    \centering
    \includegraphics[width=\linewidth]{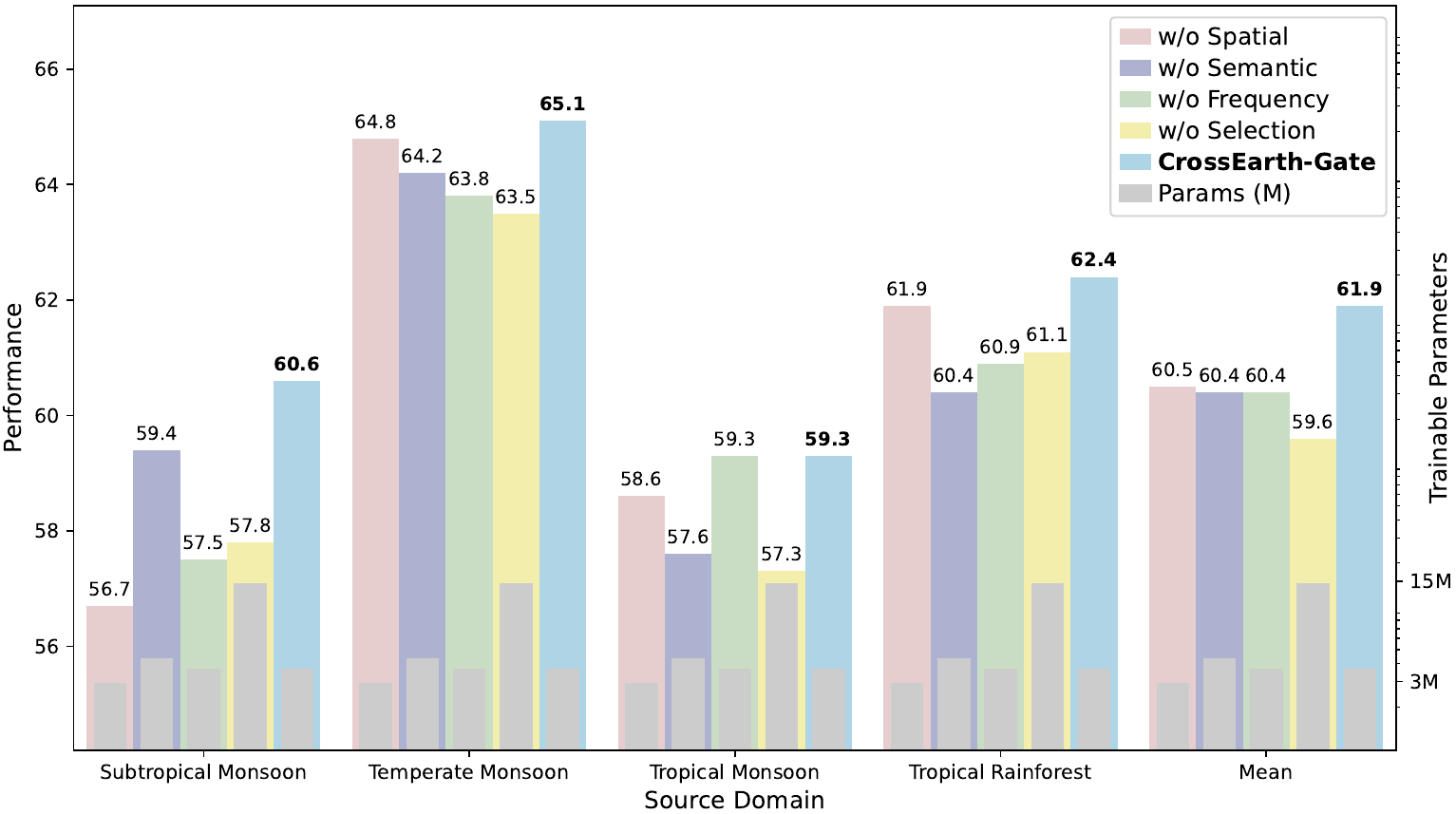}
    \caption{
    Ablation study of model component on CASID benchmarks. 
    We compare the performance and trainable parameters of CrossEarth-Gate against versions with key components removed.
    }
    \label{fig:ablation}
\end{figure}

\paragraph{Core Components Impact.}
We further dissect the framework by ablating its key components, with results visualized in Fig. \ref{fig:ablation}.
When all modules from the toolbox are trained simultaneously without selection, it results in the most performance degradation. 
Its failure proves that naively increasing trainable parameters is detrimental, likely leading to conflicting gradient updates. 
This result empirically validates the necessity of CrossEarth-Gate to guide the gradient flow for utilizing adaptation resources effectively.
Removing each specialized module type causes a comparable drop in mean performance.
All three adaptation pathways are non-redundant and essential for achieving comprehensive generalization. 
For instance, the Sub and Tem domains suffer most from the removal of spatial and frequency modules, while Trf is most impacted by the loss of semantic modules, underscoring the varied nature of these domain challenges.

\vspace{-10pt}
\subsection{Qualitative Comparison}
We also provide a qualitative comparison in Fig. \ref{fig:vis}, visualizing the predictions from our method, LoRA, AdaptFormer, and Earth-Adapter across six DG scenarios.
The PEFT baselines, while effective for their intended challenge, fail outside their specialization.
For example, the semantic-focused AdaptFormer fails to detect large areas of water with frequency artifacts in the Sub2Tms task. 
The spatial-related LoRA misclassifies the forest as water in the Tms2Sub task, resulting in semantic errors.
The frequency-focused Earth-Adapter demonstrates spatial discontinuity of the road in the Tem2Trf task.
Notably, CrossEarth-Gate avoids the catastrophic semantic errors, correctly captures the spatial extent of objects, and produces clean, accurate maps, mitigating the influence of artifacts.
Our method consistently produces segmentation maps that are closer to the ground-truth labels across all six scenarios, demonstrating a more robust and superior generalization capability.

%% file: sec/tab/tab1.tex
\begin{table*}[t]
\renewcommand{\arraystretch}{0.88} 
\caption{
Performance comparison on CASID benchmarks across 12 DG experiments. 
We compare specialized DA models, FMs, PEFT methods, and CrossEarth-Gate. The \textbf{best} and \underline{second-best} scores are indicated in bold and underlined, and the improvement of CrossEarth-Gate is shown in brackets.
Sub: Subtropical Monsoon. Tem: Temperate Monsoon. Tms: Tropical Monsoon. Trf: Tropical Rainforest.}
\vspace{-6pt}
\centering
\resizebox{\textwidth}{!}{
\begin{tabular}{ccc|cccc|cccc}
\toprule
Backbone & Method &  Params (M) & Sub2Tem & Sub2Tms & Sub2Trf &Average & Tem2Sub & Tem2Tms & Tem2Trf & Average\\ 
\midrule
 
\multirow{2}{*}{MiT-B5 \cite{xie2021segformer}} 
& HRDA \cite{hoyer2022hrda}  & 81.4 & 34.6 & 63.0 & 56.2 &51.3 & 62.7 & 63.3 & 44.9 &57.0 \\
& DAFormer \cite{hoyer2022daformer}  & 81.4 & 36.0 & {63.5} & 59.0 &52.8 & 59.8 & {63.8} & 51.7 &58.4 \\
\midrule

\multirow{4}{*}{ViT-L \cite{dosovitskiy2020image}} 
& MTP \cite{wang2024mtp} &304.2 & 39.4 & 60.0 & 61.6 &53.7 & {66.2} & {64.7} & 48.3 &59.7 \\
&  SatMAE \cite{cong2022satmae} &304.2  & 33.1 & 52.9 & 48.7 &44.9 & 51.8 & 43.7 & 36.2 &43.9\\
& ScaleMAE \cite{reed2023scale} &304.2  & 39.2 & 60.7 & 56.4 &52.1 & 62.5 & 57.3 & 45.9 &55.2\\
& RemoteCLIP \cite{liu2024remoteclip} &304.2 & 8.5 & 15.1 & 19.7 &14.4  & 54.3 & 57.8 & 16.7 &42.9\\
\midrule

\multirow{12}{*}{DINOv2-L \cite{oquab2023dinov2}} 
& Frozen & 0.0 &43.1 &63.4 &58.8 &55.2 &66.9 &64.8 &59.0 &63.6  \\
& Full-Tuning &304.2 &38.6 &62.3 &58.3 &53.0 &58.7 &55.8 &43.8 &52.8 \\
& CrossEarth \cite{gong2024crossearth} &3.0  & {48.1} & \underline{64.6} & \underline{64.2} &\underline{59.0}  & 63.5  & 61.8  & {57.8} &61.0 \\ 
& VPT \cite{jia2022visual} & 3.1 &41.2 &61.4 &60.8 &54.4 &64.6 &61.2 &58.7 &61.5   \\
& SLR \cite{scheibenreif2024parameter} & 6.9 &24.3 &37.0 &32.3 &31.2 &33.3 &22.0 &26.1 &27.1  \\
& Rein \cite{wei2024stronger} & 3.0 &42.3 &56.2 &52.9 &50.5 &66.0 &62.1 &56.7 &61.6   \\
& LoRA \cite{hu2022lora}  & 6.4 &\underline{48.7} &60.5 &62.3 &57.2 &\underline{68.2} &\underline{65.4} &58.2 &64.0   \\
& Adapter \cite{houlsby2019parameter} & 6.3 &46.9 &61.5 &63.3 &57.2 &67.7 &61.9 &58.5 &62.7  \\
& Adaptformer \cite{chen2022adaptformer} &3.2 &47.9 &58.4 &62.1 &56.1 &\underline{68.2} &64.4 &\underline{59.8} &64.1 \\
& Earth-Adapter \cite{hu2025earth} &9.6 &48.3 &61.6 &60.5 &56.8 &\textbf{68.8} &64.9 &\underline{59.8} &\underline{64.5} \\
\rowcolor{lightgreen}
\cellcolor{white} &\textbf{CrossEarth-Gate} &3.0-4.4   &\textbf{50.1 (+1.4)} &\textbf{66.6 (+2.0)} &\textbf{65.2 (+1.0)} &\textbf{60.6 (+1.6)} &68.0 (-0.8) &\textbf{67.0 (+1.6)} &\textbf{60.3 (+0.5)} &\textbf{65.1 (+0.6)}  \\
 
\midrule

Backbone & Method &  Params (M) &Tms2Sub  &Tms2Tem &Tms2Trf &Average & Trf2Sub & Trf2Tem & Trf2Tms & Average\\ 
 \midrule

\multirow{2}{*}{MiT-B5 \cite{xie2021segformer}} 
& HRDA \cite{hoyer2022hrda} & 81.4 & 63.8 & 33.3 & 56.9 &51.3 & 63.3 & {43.0} & 63.1 &56.5\\
& DAFormer \cite{hoyer2022daformer} & 81.4 & 61.8 & 31.4 & 56.2 &49.8 & 62.9 & 39.6 & 62.7 &55.1\\
\midrule
 
\multirow{4}{*}{ViT-L \cite{dosovitskiy2020image}}  
& MTP \cite{wang2024mtp} &304.2  & 56.1 & 32.3 & {60.1} &49.5 & 55.6 & 37.0 & 59.8 &50.8\\
& SatMAE \cite{cong2022satmae} &304.2 & 60.2 & 32.5 & 53.0 &48.6  & 59.1 & 32.8 & 56.7 &49.5\\
& ScaleMAE \cite{reed2023scale} &304.2 & 62.1 & 35.2 & 55.4 &50.9 & 60.6 & 30.9 & 58.0 &49.8\\
& RemoteCLIP \cite{liu2024remoteclip} &304.2 & 22.5 & 12.3 & 22.5 &19.1 & 32.1 & 26.0 & 23.6 &27.2 \\
\midrule

\multirow{12}{*}{DINOv2-L \cite{oquab2023dinov2}} 
& Frozen & 0.0 &65.3 &37.0 &60.9 &54.4 &68.0 &43.0 &66.7 &59.2  \\
& Full-Tuning &304.2 &60.9 &29.9 &58.9 &49.9 &61.9 &32.9 &64.3 &53.0 \\
& CrossEarth \cite{gong2024crossearth} &3.0 &69.1 &40.5 &60.7 &56.8 &67.9 &42.3  &64.4 &58.2 \\ 
& VPT \cite{jia2022visual} & 3.1 &64.8 &33.9 &56.8 &51.8 &67.3 &38.0 &66.9 &57.4   \\
& SLR \cite{scheibenreif2024parameter} & 6.9 &39.6 &23.6 &38.7 &34.0 &42.2 &24.2 &34.3 &33.5  \\
& Rein \cite{wei2024stronger} & 3.0 &58.2 &29.0 &53.4 &46.9 &62.6 &33.8 &61.1 &52.5   \\
& LoRA \cite{hu2022lora}  & 6.4 &66.7 &45.2 &60.2 &54.5 &67.7 &46.4 &65.3 &59.8   \\
& Adapter \cite{houlsby2019parameter} & 6.3 &67.8  &38.5 &60.4 &55.5 &68.9 &47.7 &\underline{67.1} &61.2  \\
& Adaptformer \cite{chen2022adaptformer} &3.2 &67.9 &\underline{45.7} &60.5 &\underline{58.0} &\textbf{69.8} &\underline{48.7} &66.8 &\underline{61.8} \\
& Earth-Adapter \cite{hu2025earth} &9.6 &\underline{68.0} &39.2 &\underline{61.3} &56.2 &68.4 &47.8 &65.7 &60.6 \\
\rowcolor{lightgreen}
 \cellcolor{white} &\textbf{CrossEarth-Gate} &3.0-4.4  &\textbf{68.3 (+0.3)} &\textbf{46.3 (+0.6)} &\textbf{63.3 (+2.0)} &\textbf{59.3 (+1.3)} &\underline{69.0 (-0.8)} &\textbf{50.0 (+1.3)} &\textbf{68.1 (+1.0)} &\textbf{62.4 (+0.6)}  \\
\bottomrule
\label{exp:casid}
\end{tabular}}
\vspace{-10pt}
\end{table*}

%% file: sec/tab/tab2.tex
\begin{table*}[t]
\renewcommand{\arraystretch}{0.92} 
\caption{
Performance comparison on P(r)2Res and P(i)2Res DG experiments.  
Surf: Impervious surfaces. Bldg: Building. Clut: Clutter.}
\vspace{-3pt}
\centering
\resizebox{\textwidth}{!}{
\begin{tabular}{ccccccccccccccc}
\toprule
\multirow{2}{*}{Backbone} & \multirow{2}{*}{Method} & \multirow{2}{*}{\begin{tabular}[c]{@{}c@{}}Params \\  (M)\end{tabular}} & \multicolumn{5}{c}{P(r)2Res (Classes)} & \multirow{2}{*}{mIoU (\%)} & \multicolumn{5}{c}{P(i)2Res (Classes)} & \multirow{2}{*}{mIoU (\%)} 
\\ \cline{4-8} \cline{10-14}
& & & {Surf} &{Bldg}
& {Tree} & {Car} & {Clut}& & {Surf} &{Bldg}
&{Tree}&{Car}&{Clut}\\ \midrule 

\multirow{2}{*}{MiT-B5 \cite{xie2021segformer}} 
& HRDA \cite{hoyer2022hrda}&81.4 & 28.8 &27.2 &{56.8} &4.0&66.4&36.6 &26.1&29.2&44.3&5.2&61.6&33.3\\
& DAFormer \cite{hoyer2022daformer}&81.4 & {33.0} & {32.6} &{55.0} &7.6 &{66.7}& {39.0} &34.1 &43.6 &41.7 &12.0 &58.9 & 38.1\\
\midrule 

\multirow{5}{*}{ViT-L \cite{dosovitskiy2020image}} 
&MTP \cite{wang2024mtp} &304.2 &34.3 &40.9&41.7&{16.3}&62.7&39.2 &31.3&36.8&0.5&{14.4}&57.5&28.1 \\
&{SatMAE} \cite{cong2022satmae}&304.2  &13.7 & 13.8 & 26.1 & 3.7 & 27.1 & 16.9 & 15.5 & 13.7 &24.4& 3.0&21.1& 15.5 \\

&ScaleMAE \cite{reed2023scale}&304.2  & 23.5 & 31.9 & 45.4 & 0.1 & 47.7 & 29.7 & 13.9 & 35.0 & {48.7} & 0.3 & 49.4 & 29.4\\

&RemoteCLIP \cite{liu2024remoteclip}&304.2 & 2.2 & 8.2 & 2.6 & 0.0 & 0.0 & 2.6 & 1.3 & 10.1 & 27.5 & 0.2 &0.8& 8.0 \\ 
\midrule 

\multirow{13}{*}{DINOv2-L \cite{oquab2023dinov2}} 
& Frozen & 0.0 &45.0 &\textbf{61.7} &64.1 &33.8 &66.3 &54.2 &53.0 &61.4 &56.5 &31.1 &69.1 &54.2  \\
& Full-Tuning &304.2 &21.3 &22.5 &25.2 &3.0 &61.8 &26.7 &17.5 &29.6 &64.5 &3.2 &65.3 &36.0 \\
&CrossEarth \cite{gong2024crossearth} &3.0 & {43.6}& {59.7 }& {51.0} & {16.4}& {63.2} &{46.8} &{41.9} &{60.6} & {51.3}& 10.6&{62.9} &{45.5} \\
& VPT \cite{jia2022visual} & 3.1 &50.3 &56.6 &58.8 &38.1 &72.0 &55.1 &52.8 &61.0 &58.8 &35.6 &71.1 &55.9   \\
& SLR \cite{scheibenreif2024parameter} & 6.9 &52.1 &59.3 &61.0 &21.7 &72.8 &53.4 &54.4 &57.4 &55.5 &19.3 &71.3 &51.6  \\
& Rein \cite{wei2024stronger} & 3.0 &48.8 &59.9 &51.4 &23.0 &69.4 &50.5 &21.1 &42.0 &56.3 &19.9 &59.7 &39.8   \\
& LoRA \cite{hu2022lora}  & 6.4 &51.1 &57.4 &62.5 &39.1 &72.8 &56.6 &50.1 &59.8 &\textbf{65.1} &\underline{35.1} &72.3 &56.5   \\
& Adapter \cite{houlsby2019parameter} & 6.3 &56.7 &59.6 &53.1 &30.3 &69.8 &53.9 &\underline{56.1} &59.3 &63.5 &30.7 &\underline{73.3} &56.1  \\
& Adaptformer \cite{chen2022adaptformer} &3.2 &\underline{58.5} &60.3 &\underline{65.5} &38.9 &\underline{73.1} &\underline{59.2} &55.4 &\textbf{61.9} &60.2 &\textbf{37.1} &72.6 &\underline{57.4} \\
& Earth-Adapter \cite{hu2025earth} &9.6 &56.5 &\underline{61.1} &63.4 &\underline{39.6} &72.4 &58.6 &54.2 &60.8 &61.8 &\underline{35.1} &72.8 &57.0  \\
\rowcolor{lightgreen}
 \cellcolor{white}&   & &\textbf{60.8} &58.6 &\textbf{67.5} &\textbf{40.5} &\textbf{73.3} &\textbf{60.1} &\textbf{57.7} &\underline{61.5} &\underline{64.6} &34.6 &\textbf{74.7} &\textbf{58.6}
 \\
\rowcolor{lightgreen}
 \cellcolor{white}& \multirow{-2}{*}{\textbf{CrossEarth-Gate}} &  \multirow{-2}{*}{3.3-4.0}&\textbf{(+2.3)} &(-3.1) & \textbf{(+2.0)} &\textbf{(+0.9)} &\textbf{(+0.2)} &\textbf{(+0.9)}
 &\textbf{(+1.6)} &\underline{(-0.4)} & \underline{(-0.5)} &(-2.5) &\textbf{(+1.4)} & \textbf{(+1.2)}
 \\

\bottomrule
\end{tabular}}

\label{exp:rescue}
\end{table*}

%% file: sec/tab/tab3.tex
\begin{table}[t]
\renewcommand{\arraystretch}{0.95} 
\caption{Performance comparison on DA 
benchmarks between representative DA models, PEFT methods, and CrossEarth-Gate.}
 \resizebox{\linewidth}{!}{
\begin{tabular}{ccccccc}
\toprule

\multirow{2}{*}{Methods} & \multirow{2}{*}{\begin{tabular}[c]{@{}c@{}}Params \\ (M) \end{tabular}} & \multicolumn{4}{c}{Domain Adaptation} &\multirow{2}{*}{Avg.} \\
\cline{3-6}
 &   &P2V &V2P &R2U &U2R
 \\
\midrule

HRDA \cite{hoyer2022hrda} &81.4  & 67.6  & 58.6 & 53.2 & 35.3 & 53.7 \\
DAFormer \cite{hoyer2022daformer} & 81.4  & 64.4  & 54.8 & 52.7 & 42.5 & 53.6  \\
\midrule
 Frozen & 0.0 &66.1  &59.2  & 54.9 & 45.5 &56.4  \\
 Full-Tuning &304.2 & 62.4 & 59.6 &42.6  &35.8  &50.1  \\
 VPT \cite{jia2022visual} & 3.1 & 65.0  & 57.6  &55.8  &47.0  &56.4  \\
 SLR \cite{scheibenreif2024parameter} & 6.9 &14.4  &16.5 &23.8  &9.6  &16.1  \\
 Rein \cite{wei2024stronger} & 3.0 &64.1  & 59.5 &54.7  &34.5  &53.2  \\
 LoRA \cite{hu2022lora}  & 6.4 & 65.2  &\textbf{62.4}  &\underline{56.2}  &46.6  &57.6  \\
 Adapter \cite{houlsby2019parameter} & 6.3 & 66.8 &60.3  &53.6  &47.7  &57.1  \\
 Adaptformer \cite{chen2022adaptformer} &3.2 &66.9 &62.2 &53.7 &\underline{48.1} & 57.7\\
 Earth-Adapter \cite{hu2025earth} &1.0-3.9 &\underline{67.1}  &61.6  &56.0  &47.5  &\underline{58.1} \\
\rowcolor{lightgreen}
  &  & \textbf{68.2} &\textbf{62.4}  &\textbf{56.5}  &\textbf{49.1} &\textbf{59.1} \\

 \rowcolor{lightgreen}
  \multirow{-2}{*}{\textbf{CrossEarth-Gate}} & \multirow{-2}{*}{1.7-3.9} & \textbf{(+1.1)} & \textbf{(+0.0)} & \textbf{(+0.3}) &\textbf{(+1.0)} &\textbf{(+1.0)}
  \\ 
\bottomrule
\end{tabular}
}
\centering
\label{exp:da}
\end{table}

%% file: sec/tab/tab4.tex
\begin{table}[t]
\renewcommand{\arraystretch}{0.94} 
\caption{
Ablation studies of model backbone on CASID benchmarks. 
We only show the average performance of four climate domains as the source domain, respectively.
We demonstrate the generalizability of CrossEarth-Gate across different backbones.
}
    \centering
    \setlength\tabcolsep{2.5pt} 
    \resizebox{\linewidth}{!}{
\begin{tabular}{cccccccc}
\toprule
\multirow{2}{*}{Backbone} & \multirow{2}{*}{Method} & \multirow{2}{*}{\begin{tabular}[c]{@{}c@{}}Params \\ (M) \end{tabular}}  
& \multicolumn{4}{c}{Source Domain} & \multirow{2}{*}{Mean} \\
\cline{4-7}
 & & & Sub &Tem & Tms & Trf  \\ 

\midrule

\multirow{3}{*}{
{\begin{tabular}[c]{@{}c@{}} SatMAE \\ (Large) \cite{cong2022satmae} \end{tabular}}
} 
& Frozen &0.0 &24.4 &21.5  &24.8 &26.3 &24.3\\
&  Full-Tuning &304.2  &35.4 &14.9 &35.2 &29.8 &28.8\\
&\cellcolor{lightgreen}\textbf{CrossEarth-Gate}  &\cellcolor{lightgreen}2.9-4.0  &\cellcolor{lightgreen}\textbf{37.1} &\cellcolor{lightgreen}\textbf{21.8} &\cellcolor{lightgreen}\textbf{36.8} &\cellcolor{lightgreen}\textbf{31.8} &\cellcolor{lightgreen}\textbf{31.9}\\
\midrule

\multirow{3}{*}{
{\begin{tabular}[c]{@{}c@{}} Scale-MAE \\ (Large) \cite{reed2023scale} \end{tabular}}
} 
& Frozen &0.0 &50.7  &48.5  &50.9 &50.9 &50.3\\
&  Full-Tuning &304.2  &52.8 &57.6 &51.9 &52.6 &53.7\\
&\cellcolor{lightgreen} \textbf{CrossEarth-Gate}  &\cellcolor{lightgreen} 2.9-3.6  &\cellcolor{lightgreen}\textbf{56.3} &\cellcolor{lightgreen}\textbf{58.1} &\cellcolor{lightgreen}\textbf{54.9} &\cellcolor{lightgreen}\textbf{53.8} &\cellcolor{lightgreen}\textbf{55.8}\\
\midrule

\multirow{3}{*}{
{\begin{tabular}[c]{@{}c@{}} SAM \\(Huge) \cite{kirillov2023segment}\end{tabular}}
} 
& Frozen                    &0.0   &53.3  &57.5  &53.2 &53.4 &54.4\\
&  Full-Tuning              &631.2 &54.7  &58.2  &51.7 &54.6 &54.8\\
&\cellcolor{lightgreen}\textbf{CrossEarth-Gate} &\cellcolor{lightgreen}4.0-4.9       &\cellcolor{lightgreen}\textbf{56.1} &\cellcolor{lightgreen}\textbf{61.9} &\cellcolor{lightgreen}\textbf{56.3} &\cellcolor{lightgreen}\textbf{59.3} &\cellcolor{lightgreen}\textbf{58.4} \\
\midrule

\multirow{3}{*}{
{\begin{tabular}[c]{@{}c@{}} DINOv2\\(Small)  \cite{oquab2023dinov2}\end{tabular}}
} 
 & Frozen      &0.0   &49.5  &59.6 &51.9  &52.7  &53.4\\
&  Full-Tuning &22.1  &49.4 &49.3  &48.7  &50.2  &49.4\\

 &\cellcolor{lightgreen}\textbf{CrossEarth-Gate}  &\cellcolor{lightgreen}0.7  &\cellcolor{lightgreen}\textbf{54.6} &\cellcolor{lightgreen}\textbf{62.0} &\cellcolor{lightgreen}\textbf{52.7} &\cellcolor{lightgreen}\textbf{55.8} &\cellcolor{lightgreen}\textbf{56.2}\\
\midrule

\multirow{3}{*}{
{\begin{tabular}[c]{@{}c@{}} DINOv2 \\(Base) \cite{oquab2023dinov2}\end{tabular}}
} 
& Frozen                    &0.0        &51.3 &62.5 &52.8 &54.4 &55.3 \\
& Full-Tuning               &86.6       &50.2 &53.7 &50.2 &49.4 &50.9\\
& \cellcolor{lightgreen}\textbf{CrossEarth-Gate}  &\cellcolor{lightgreen}1.4-1.8    & \cellcolor{lightgreen}\textbf{55.6} &\cellcolor{lightgreen}\textbf{63.6} &\cellcolor{lightgreen}\textbf{54.7} &\cellcolor{lightgreen}\textbf{57.4} &\cellcolor{lightgreen}\textbf{57.8}\\

\bottomrule
\label{exp:ablation}
\end{tabular}}
\end{table}

%% file: sec/5_conclusion.tex
\section{Conclusion}
This paper addresses the limitations of current static, specialized PEFT methods in cross-domain RS adaptation. 
Existing approaches often fall short by focusing on a single facet of the complex spatial, semantic, and frequency shifts inherent to RS data. 
We introduce CrossEarth-Gate to tackle this limitation through two key innovations: a comprehensive RS module toolbox and a Fisher-guided adaptive selection mechanism. 
The toolbox integrates three module types to provide distinct functional pathways of models. 
Subsequently, by leveraging Fisher Information to quantify module importance, CrossEarth-Gate dynamically directs the gradient flow, activating only the most critical modules for the specific adaptation task.
Comprehensive experiments demonstrate the effectiveness, efficiency, generalizability, and explainability of our method.

%% file: sec/6_suppl.tex
\setcounter{page}{1}
\onecolumn
\begin{center}
        \Large
        \textbf{\thetitle}\\
        \vspace{0.5em}Appendix \\
        \vspace{1.0em}
\end{center}
\appendix

\renewcommand{\contentsname}{}
\startcontents[sections]
\printcontents[sections]{}{1}{\section*{Contents}\setcounter{tocdepth}{3}} 
\newpage

\section{Overview}
This Appendix provides additional details, in-depth analysis, and implementation specifics. 
In Sec. \ref{discuss}, we elaborate on the trade-offs regarding storage and computational costs, clarify the motivation behind the unified RS module toolbox, and discuss potential avenues for future generalization.
In Sec. \ref{studies}, we present a dynamic network analysis to visualize the explainability of the Fisher-guided mechanism, followed by complete ablation studies across various backbones, module complementarity analysis, computational analysis, and extensive qualitative comparisons of segmentation results.
In Sec. \ref{details}, we provide rigorous descriptions of the cross-domain benchmarks (CASID, ISPRS, LoveDA, RescueNet), detailed implementation specifications for the eight PEFT baselines, and the exact hyperparameter configurations for CrossEarth-Gate.

\section{Discussion}
\label{discuss}
\subsection{Storage and Computing}
The Fisher-guided adaptive selection mechanism of CrossEarth-Gate dynamically activates specific, lightweight modules from the Remote Sensing (RS) module toolbox during training to optimally address the multifaceted challenges, \ie, spatial, semantic, and frequency shifts, prevalent in RS cross-domain adaptation. 
As our experiments demonstrate, this selective guidance of the gradient flow not only enhances parameter efficiency but also yields superior adaptation performance by preventing the overfitting seen in full fine-tuning and the performance degradation from naively training all modules at once.

However, this dynamic framework introduces a noteworthy consideration regarding the cumulative storage and computational costs.
During training, the framework is highly efficient as it only activates and computes gradients for a subset of Top-k modules at any given iteration. However, as the adaptation progresses, the set of modules selected at different stages may expand.
A ``worst-case'' scenario arises if the model, in its effort to adapt, eventually selects and updates all modules in the toolbox at least once. 
This would necessitate storing the parameters for the entire toolbox to constitute the final adapted model.
Consequently, at inference time, the computational path would require passing the features through all of these activated modules.
Fortunately, a core design principle of the RS module toolbox is its lightweight nature. 
Even in this hypothetical worst-case scenario, the total parameter overhead remains minimal. 
For our experiments, the cumulative size of the entire toolbox is approximately 14.4M. 
When contrasted with the 304.2M parameters of the frozen DINOv2-L \cite{oquab2023dinov2} backbone, this overhead constitutes less than 5\% of the backbone's size. 
This modest cost is an acceptable tradeoff for the significant and consistent performance gains and robust generalization achieved across 16 benchmarks.

Therefore, this opens avenues for future exploration of this tradeoff. 
For instance, a post-hoc pruning strategy could be implemented to permanently discard modules that received consistently low Fisher importance scores, thereby creating a more compact final model for inference.
One could also investigate parameter-sharing techniques where a single, lightweight module (\eg, one spatial adapter) is shared across multiple layers, with its activation still governed by the layer-specific Fisher-guided gate. 
This would further reduce the total storage footprint without sacrificing dynamic adaptation.

\subsection{Toolbox Modules}
A foundational observation of this paper is the failure of existing specialized Parameter-Efficient Fine-Tuning (PEFT) methods when applied to cross-domain RS adaptation. 
We identify that RS domain gaps are uniquely multifaceted, manifesting as a complex interplay of spatial shifts (\eg, changes in object scale and structure), semantic shifts (\eg, different class appearances), and frequency shifts (\eg, textural noise or sensor-based artifacts).
The core limitation of prior work is its ``single-pathway'' design. 
As illustrated in our analysis, methods like LoRA \cite{hu2022lora} (spatial), AdaptFormer \cite{chen2022adaptformer} (semantic), and Earth-Adapter \cite{hu2025earth} (frequency) specialize in one functional pathway. 
This specialization is their critical weakness: each method may capture only one facet of the RS domain shifts while leaving others unaddressed. 
This results in the specific, catastrophic failures we observed, such as LoRA misclassifying high-frequency waves (a frequency-domain failure) or AdaptFormer shattering the spatial continuity of a road (a spatial-domain failure).
This clear gap motivates the design of our RS module toolbox. 
Instead of proposing yet another specialized module, we establish a unified framework to comprehensively tackle these intertwined challenges by integrating modules that target each of the three distinct functional pathways. 
In this work, we instantiate the toolbox using representative PEFTs with spatial, semantic, and frequency models. 

We acknowledge that this implementation, while proven effective, is a first step. 
The challenges in RS are vast, and other types of domain shifts (\eg, temporal, atmospheric, or sensor-specific variations) may exist that are not fully captured by our current spatial-semantic-frequency-trichotomy.
This provides two clear avenues for future work.
Future research could identify and model these other potential domain gaps, leading to the development and integration of new module types into the toolbox.
Moreover, the specific methods chosen to implement each module (LoRA, Adapter, Earth-Adapter) could be substituted. 
One could explore other reparameterization PEFTs for the spatial module or different additive or prompt-based methods for the semantic module, potentially offering different performance-efficiency tradeoffs.

\subsection{Generalization}
In this paper, we have comprehensively validated the effectiveness of CrossEarth-Gate on the task of cross-domain semantic segmentation in RS. 
A key strength demonstrated in our ablations is the framework's versatility across several State-Of-The-Art (SOTA) Transformer-based \cite{vaswani2017attention} foundation models, including the DINOv2 series \cite{oquab2023dinov2}, SAM \cite{kirillov2023segment}, SatMAE \cite{cong2022satmae}, and Scale-MAE \cite{reed2023scale}. 
Our method consistently outperforms baselines, confirming its robustness within this architectural class. 
However, we acknowledge two primary avenues for exploring the broader generalizability of our approach. 
Our current validation is confined to Transformer-based backbones. 
While these represent the SOTA, a significant direction for future work is to investigate the applicability of CrossEarth-Gate to other foundational architectures, such as modern Convolutional Neural Networks. 
This would test whether our core methodology is a universal adaptation strategy or one uniquely suited to the functional pathways (\eg, Multi-Head Self-Attention (MSA) and Multi-Layer Perception (MLP) layers) of Transformers.
Moreover, the multifaceted domain gaps we identify (spatial, semantic, and frequency) are fundamental challenges in RS, extending far beyond semantic segmentation. 
Our experiments focus on this single task. 
A valuable and logical next step is to apply and evaluate the CrossEarth-Gate framework on other critical cross-domain RS tasks, such as object detection, change detection, or land-cover classification. 
This would provide a more rigorous test of our framework's robustness and further confirm the hypothesis that dynamically addressing these three-domain shifts is essential for effective RS adaptation.

\begin{figure*}[t]
    \centering

    \begin{subfigure}[b]{\textwidth}
        \centering
        \includegraphics[width=\linewidth]{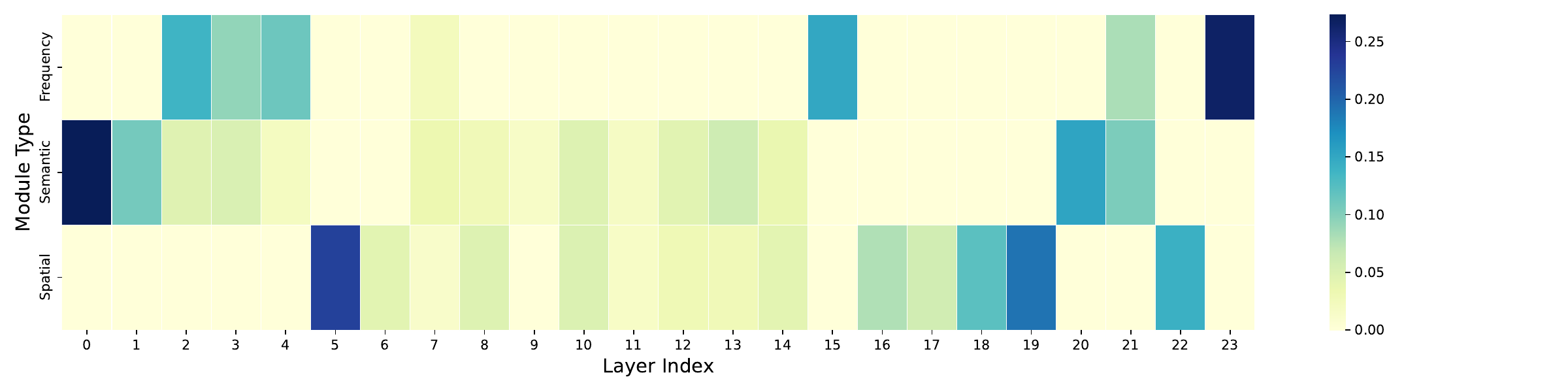}
        \caption{Aggregated importance intensity of different module types across layers of Tem.}
        \label{fig:heatmap}
    \end{subfigure}

    \begin{subfigure}[b]{\textwidth}
        \centering
        \includegraphics[width=\linewidth]{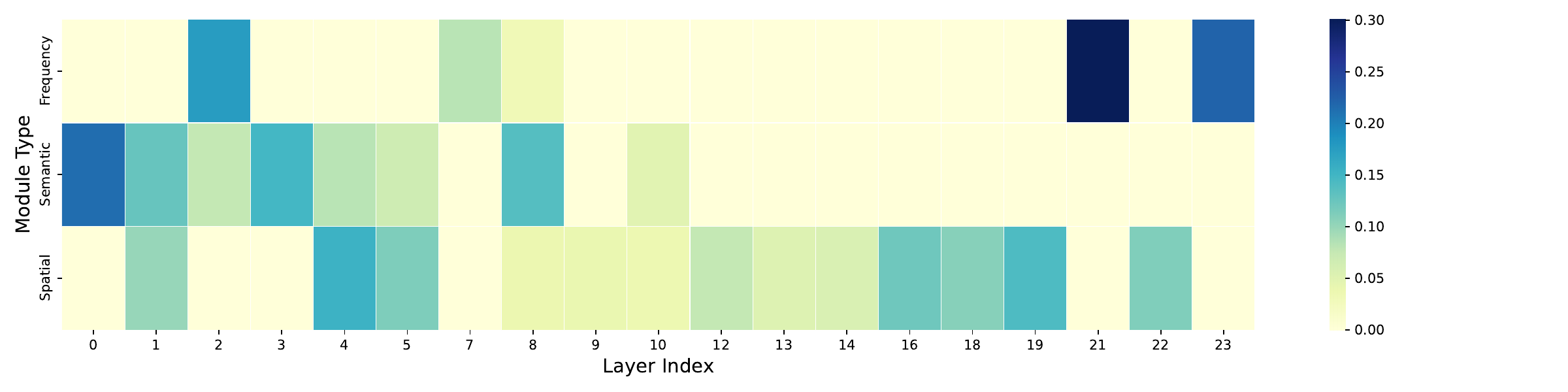}
        \caption{Aggregated importance intensity of different module types across layers of Sub.}
        \label{fig:heatmap1}
    \end{subfigure}

    \begin{subfigure}[b]{\textwidth}
        \centering
        \includegraphics[width=\linewidth]{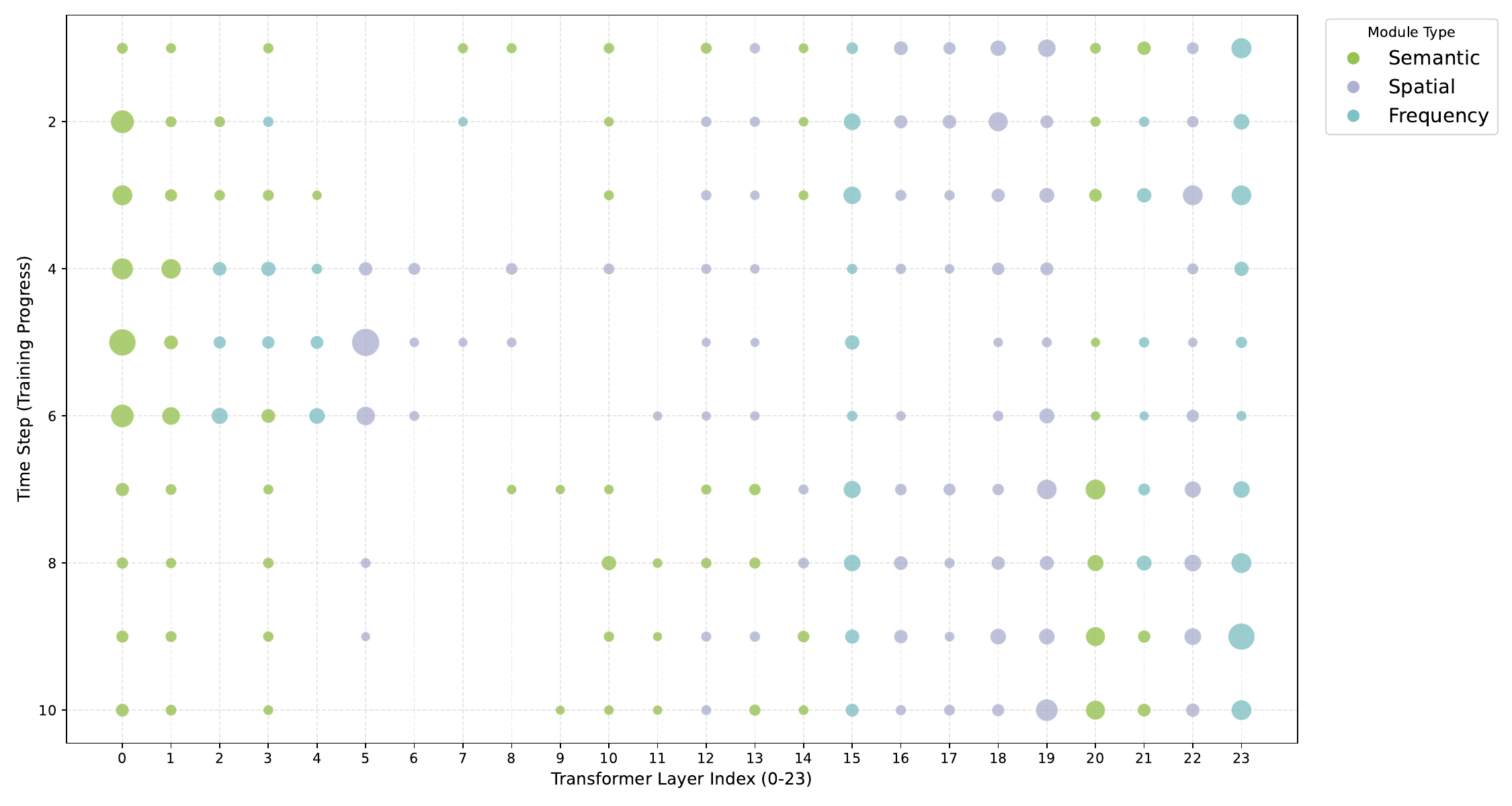}
        \caption{The evolution of module selection and importance during training of Tem.}
        \label{fig:bubble}
    \end{subfigure}

    \caption{Dynamic network analysis. 
    (a) The vertical axis represents training steps (flowing downwards), and the bubble size corresponds to importance score on the Tem.
    (b) Aggregated importance intensity of different module types across layers in the entire process on the Tem. 
    (C) Aggregated importance intensity of different module types across layers in the entire process on the Sub. 
    } 
    \label{fig:dynamic_analysis}
\end{figure*}

\section{Additional Studies}
\label{studies}

\subsection{Dynamic Network Analysis}

To further investigate the explainability of our proposed Fisher-guided adaptive selection mechanism, we conduct an in-depth dynamic network analysis of the Temperate Monsoon (Tem) domain generalization experiment, as illustrated in Fig. \ref{fig:heatmap} and \ref{fig:bubble}. 
Specifically, we record the evolution of module selection of each module type (Spatial, Semantic, and Frequency) across different layers of the DINOv2-L \cite{oquab2023dinov2} backbone (visualized via bubble size and distribution) and aggregated importance intensity of different module types across layers  (visualized via the heatmap).
Furthermore, we provide another intensity heatmap for the Subtropical Monsoon (Sub) domain in the Fig. \ref{fig:heatmap1}.

\subsubsection{Layer-wise Module Distribution} 
As shown in Fig. \ref{fig:heatmap}, the aggregated importance heatmap reveals a non-uniform, layer-dependent activation pattern. 
This hierarchy suggests that CrossEarth-Gate effectively decomposes the complex domain shift into distinct sub-problems, addressing them at the most appropriate network depths:

\begin{itemize} 

\item \textbf{Shallow Layers:} 
In the shallow layers of the network, the model predominantly activates Semantic Modules. 
This indicates that during the initial stages of feature extraction, the adaptation process focuses on realigning fundamental semantic concepts such as the variations in vegetation tonality caused by climatic differences. 
The efficient feature projection capabilities of Adapters play a critical role in this low-level semantic alignment.

\item \textbf{Middle Layers:} As information propagates to the middle layers, the selection shifts towards a hybrid configuration dominated by Spatial Modules.
This suggests that after modifying semantics, the model focuses on adjusting to geometric shifts. 
Given that the intermediate layers of Transformer \cite{vaswani2017attention} architectures are typically responsible for modeling long-range dependencies and spatial context, the activation of LoRA implies a targeted fine-tuning of the self-attention mechanism. 
This effectively mitigates spatial shifts inherent to the Tem domain, such as variations in object scale and geometric layout.

\item \textbf{Deep Layers:} 
A notable observation lies in the deep layers, where Frequency Modules exhibit a commanding presence. 
While deep features encapsulate highly abstract structural information, they are also most susceptible to frequency-domain artifacts arising from varying imaging conditions, such as spectral noise and textural distortion. 
The adaptive activation of frequency modules suggests the model possesses an implicit ``awareness'' of the necessity for spectral denoising and detail refinement prior to generating the final prediction.
\end{itemize}

\subsubsection{Domain-Specific Activation Patterns}
As shown in Fig. \ref{fig:heatmap} and \ref{fig:heatmap1}, the Sub triggers a higher activation of Spatial modules and fewer Semantic modules compared to the Tem domain. 
This distinct activation pattern aligns with the specific characteristics of the Sub shift. 
This comparative analysis further substantiates that CrossEarth-Gate does not rely on a static or uniform adaptation strategy. 
Instead, it dynamically tailors its module topology to the unique, multifaceted requirements of each specific domain shift.

\subsubsection{The Evolution of Training Dynamics}
The temporal evolution of module selection, as visualized in the Fig. \ref{fig:bubble}, demonstrates that CrossEarth-Gate operates as a dynamic system rather than a static ensemble.
Early in the training phase, we observe a broader distribution of active modules with fluctuating Fisher scores, indicating an ``exploration'' phase where the gradient flow probes various pathways to identify the most effective adaptation routes. 
As training progresses, the selection stabilizes into the distinct functional zones described above. 
This temporal behavior confirms that the Fisher-guided mechanism successfully acts as a ``gating'' controller, as it does not merely select parameters randomly but progressively converges on a resource-efficient configuration that maximizes the reduction of task-specific loss. 
This ability to ``learn how to adapt'' ensures that the model's capacity is not wasted on redundant layers, but is instead dynamically allocated to where the domain gap is most severe.

The training patterns observed on Tem demonstrate that CrossEarth-Gate possesses both parameter efficiency and explainable adaptive capability.
Rather than relying on a static, specialized fine-tuning paradigm, it employs a dynamic resource allocation strategy dictated by the specific hierarchical requirements of the domain shift. 
This adaptive, layer-wise governance is the core factor enabling CrossEarth-Gate to achieve SOTA performance in complex cross-domain tasks.

\subsubsection{Qualitative Gradient Saliency}

\begin{figure}[h]
    \centering
    \includegraphics[width=0.7\linewidth]{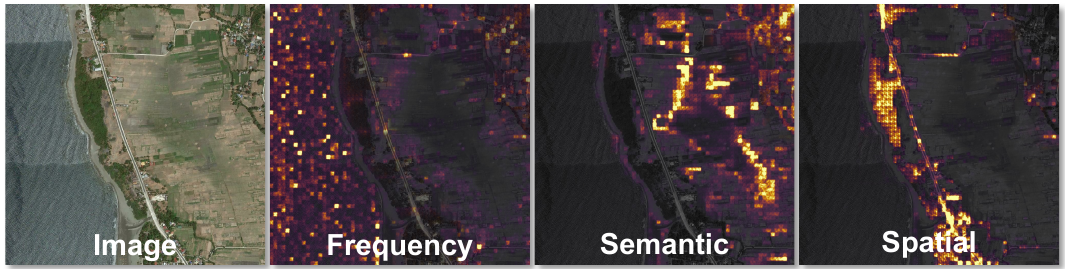}
    \caption{Visualization of gradient saliency maps for the Frequency, Semantic, and Spatial modules.}
    \label{fig:gradient}
\end{figure}

To further investigate the interpretability of our framework and explicitly link module activations to the learned visual representations, we compute gradient saliency maps for each module type with respect to the input pixels. 
As illustrated in Fig. \ref{fig:gradient}, these visualizations confirm that the Spatial, Semantic, and Frequency modules dynamically attend to distinct, functionally relevant regions of the image, corroborating our multi-faceted adaptation hypothesis. 
Specifically, the Frequency modules exhibit high activation around high-frequency image components. Conversely, the Spatial modules concentrate their gradients on prominent geometric structures. Meanwhile, the Semantic modules target broader, class-specific regions characterized by distinct visual appearances. 

\input{sec/tab/tab5}

\subsection{Complete Ablation Studies}
We provide the complete results of the ablation studies in Tab. \ref{exp:completeablation} to further evaluate the generalizability of CrossEarth-Gate across diverse foundation model architectures and to rigorously dissect the contribution of each component within our framework.
The experiments are conducted on the CASID \cite{liu2023large} dataset, covering 12 domain generalization (DG) scenarios.

\subsubsection{Generalizability Across Backbones}
We evaluate CrossEarth-Gate against ``Frozen" and ``Full-Tuning'' baselines across five distinct backbone architectures ranging from masked autoencoders \cite{he2022masked} (SatMAE \cite{cong2022satmae}, Scale-MAE \cite{reed2023scale}) to segmentation-specific models (SAM \cite{kirillov2023segment}) and self-supervised vision transformers (DINOv2 series \cite{oquab2023dinov2}). 
While it is intuitive to assume that updating all parameters would yield the highest adaptation capacity, our empirical results demonstrate a ``Full-Tuning Paradox''. 
As evidenced in the DINOv2-S and DINOv2-B experiments, Full-Tuning results in a precipitous performance decline compared to even the Frozen baseline (e.g., a drop of average performance from 59.6\% to 49.3\% average mIoU on DINOv2-S in the Tem source). 
We attribute this phenomenon to catastrophic forgetting, where unconstrained optimization on a smaller, domain-specific source dataset erodes the robust, generalizable feature representations acquired during large-scale pre-training. 
Furthermore, Full-Tuning appears highly susceptible to overfitting the specific noise patterns of the source domain, thereby hampering transferability to unseen target domains.
In contrast, CrossEarth-Gate consistently surpasses the Frozen baseline while updating less than 2\% of the parameters, effectively balancing plasticity (adaptation) and stability (generalization). 
Even on the SAM, which possesses a massive parameter count (632M) and strong zero-shot capabilities, CrossEarth-Gate improves the average mIoU by roughly 3-4\% across tasks. 
This indicates that our Fisher-guided mechanism effectively locates and tunes the sparse ``skill neurons'' required for domain alignment, even in highly capable, frozen architectures.

\subsubsection{Component Dissection}
The ablation study on the DINOv2-Large backbone validates the necessity of our unified toolbox and selection mechanism.
A critical observation is that the w/o Selection variant (activating all modules simultaneously) consistently underperforms the full CrossEarth-Gate. 
For example, in the Tem source experiments, removing selection drops the average mIoU from 65.1\% to 63.5\%. 
This empirically indicates that naively activating all adaptation pathways introduces gradient conflict and noise. The Fisher-guided selection acts as a necessary gate, ensuring that spatial, semantic, and frequency updates do not interfere with one another.
The removal of specific modules (w/o Spatial/Semantic/Frequency) leads to performance degradation, though the impact varies by domain. 
Notably, the w/o spatial variant causes the most significant drop in the Sub source scenarios. 
This suggests that the shift from Subtropical Monsoon to other zones involves significant spectral and phenological changes (e.g., vegetation density, lighting conditions) but retains relatively consistent spatial layouts and structural geometries.

\subsubsection{Analysis of Edge Cases and Limitations}
While CrossEarth-Gate achieves state-of-the-art performance in the majority of scenarios, we observe isolated cases where baselines or ablated versions show marginal superiority.
For example, in the benchmark of Tms2Sub using Dinov2-L, the variant without Spatial modules achieves a higher score (70.0\%) than the full method (68.3\%). 
The shift from Tropical Monsoon to Subtropical Monsoon involves climate zones with relatively similar object scales and geometric structures. Consequently, the domain gap may be predominantly textural (frequency) rather than geometric (spatial). 
In this context, the introduction of LoRA modules might have introduced unnecessary inductive bias, slightly disrupting the spatial integrity of the pre-trained features. As for Trf2Tem, the variant activating all modules performs best (51.0\%). 
This is a scenario with an extreme domain shift. It is possible that the domain gap is so multifaceted and severe that the model requires every available degree of freedom to adapt, outweighing the negative impact of gradient conflict. However, this comes at the cost of tripling the trainable parameters, which reduces the method's efficiency. 
Therefore, despite these isolated outliers, the CrossEarth-Gate framework remains the superior strategy, as it provides the requisite plasticity to address multifaceted shifts on average, ensuring global robustness without requiring manual, domain-specific architecture tuning.

\subsection{Module Complementarity and Synergy Analysis}
Regarding module complementarity, we further conduct a granular decomposition ablation on the Sub domain in the Tab \ref{alb}. 
The combination of all three modules outperforms single-module baselines, confirming that they address complementary domain shifts.
However, the gain from naive combination is marginal, which implies that without explicit gating, ``feature interference'' suppresses the full potential of the toolbox. 
Randomly selecting modules helps mitigate interference slightly and improves performance. 
However, the Fisher-Guided mechanism is the key to unlocking true synergy, which boosts the performance significantly. 
By activating only the most information-rich pathway, it facilitates the toolbox modules to capture complementary aspects of the domain shift.

\begin{table}[h!]
\centering
\caption{Granular decomposition ablation study on the Sub domain.}
\begin{tabular}{ccccc|cccc}
\toprule
Spatial & Semantic & Frequency &Random & Fisher & Tem &Tms &Trf &Avg. \\ 
\midrule
$\checkmark$ & & & & &48.7 &60.5 &62.3 &57.2 \\
 &$\checkmark$ & & &  &47.9 &58.4 &62.1 &56.1\\
  & &$\checkmark$ & &  
  & 47.8 &61.1 &59.2 &56.0 \\
 $\checkmark$&$\checkmark$ & & & 
 & 49.3 &60.4 &62.8 &57.5\\
 &$\checkmark$ &$\checkmark$ & & 
 & 48.5 &61.0 &62.2 &57.2\\
$\checkmark$ & &$\checkmark$ & & 
 &48.6 &61.7 &62.4 &57.6\\
$\checkmark$  &$\checkmark$ &$\checkmark$ & & &48.0 &61.8 &63.5 &57.8\\
$\checkmark$  &$\checkmark$ &$\checkmark$ &$\checkmark$ & &48.6 &64.3 &63.1 &58.7 \\
$\checkmark$  &$\checkmark$ &$\checkmark$ & &$\checkmark$ &50.1 &66.6 &65.2 &60.6 \\
  
\bottomrule
\end{tabular}

\label{alb}
\end{table}

\subsection{Additional Computational Analysis}
As detailed in Appendix \ref{details}, the selection process is performed 10 times during the 30,000 training iterations (update interval $N=3000$, calculation samples $M=100$) for CASID benchmarks.
As shown in the Tab. \ref{tab:efficiency_comparison}, we quantify the computation costs using the CASID \cite{liu2023large} benchmarks on the Sub souce domain.
The total time consumed by the selection process constitutes only $\sim$3.7\% of the training time.
For practical deployment, the Fisher-guided selection is exclusively a training-time optimization. 
Once training is complete, the module topology is frozen as other PEFT methods do.
Our inference throughput and computational cost are highly competitive with static PEFT baselines. 

\begin{table}[h]
    \centering
    \caption{Comparison of efficiency metrics including parameters, training time, memory usage, FLOPs, and throughput on the Sub domain.}

    \begin{tabular}{lccccc}
    \toprule
    \multirow{2}{*}{Method} & \multirow{2}{*}{\begin{tabular}[c]{@{}c@{}}Params \\  (M)\end{tabular}}
    & \multicolumn{2}{c}{Training}  &\multicolumn{2}{c}{Inference} 
    \\ 
    \cmidrule(lr){3-4} \cmidrule(lr){5-6}
    & & Time (min) & Mem (GB) & GFLOPs & Throughput (img/s) 
    \\
    \midrule
    Selection & 14.4 & 16  & 16.1 & N/A & N/A \\
    \textbf{Ours} & 3.0--4.4 & 437 & 12.8 & 1258 & 2.12 
    \\
    \midrule
    LoRA & 6.4 & 441 & 12.3 & 1242 & 2.18 \\
    AdaptFormer & 3.2 & 420 & 12.1 & 1255 & 2.14 \\
    Earth-Adapter & 9.6 & 470 & 15.2 & 1281 & 1.83 \\
    Full-tuning & 304.2 & 617 & 18.2 & 1242 & 2.18 \\
    \bottomrule
    \end{tabular}
    
    \label{tab:efficiency_comparison}
\end{table}

\subsection{Additional Qualitative Comparison}
In this section, we provide a more comprehensive qualitative analysis of the segmentation results. 
We present visualizations across diverse benchmarks, including the cross-climate scenarios of CASID (Sub2Tms, Tem2Trf, Tms2Sub, Trf2Tem), the disaster adaptation scenarios of RescueNet (P(r)2Res, P(i)2Res), and the domain adaptation benchmarks (P2V, V2P, R2U, U2R).
The core motivation of CrossEarth-Gate is to address the limitations of existing specialized PEFT methods, which typically focus on a single functional pathway (spatial, semantic, or frequency). 

\begin{figure*}[h]
    \centering
    \includegraphics[width=\linewidth]{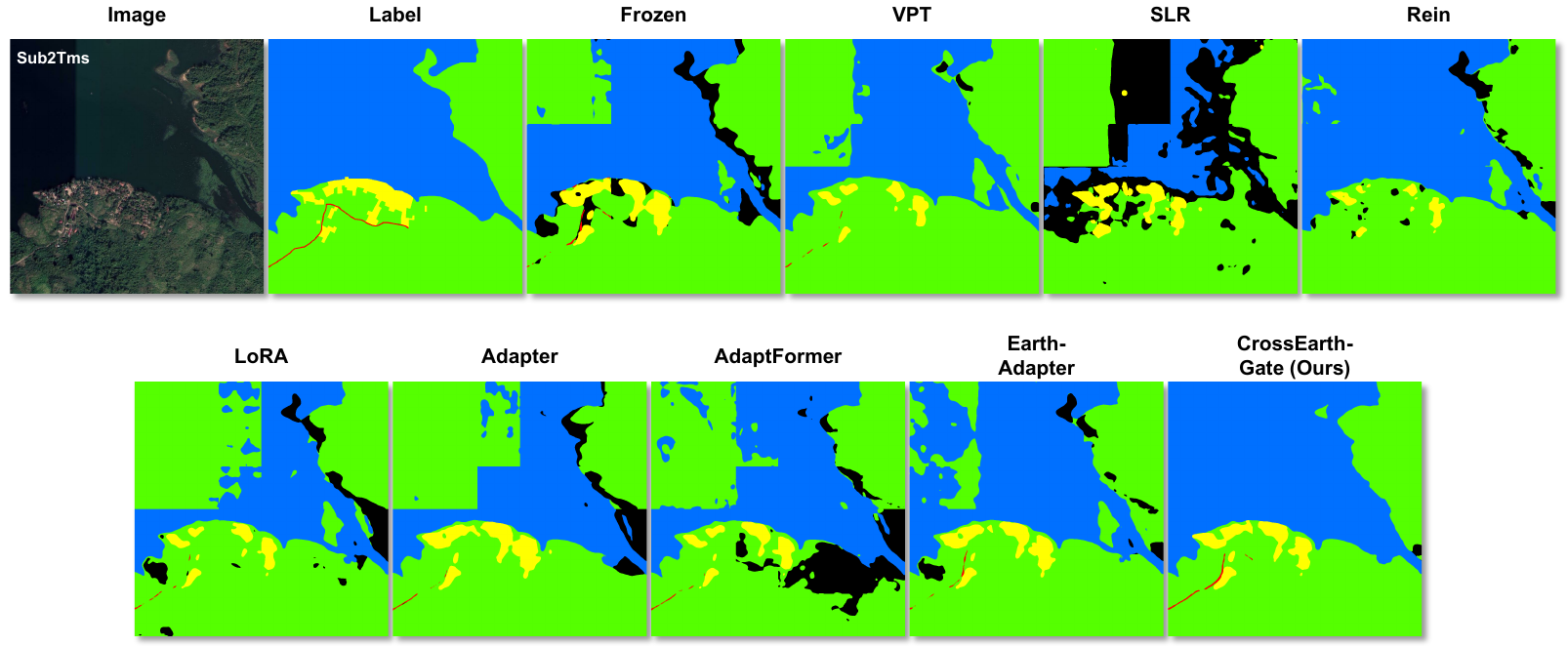}
    \caption{Complete visualizations of predicted segmentation maps of PEFT methods. 
    These samples are collected from the domain generalization benchmarks of Sub2Tms on the CASID \cite{liu2023large} dataset, where {\color{red} red} is the road class, {\color{yellow} yellow} is the building class, {\color{blue} blue} is the water class, {\color{green} green} is the forest class, and black is the background class.
    }
    \label{fig:sub2tms}
\end{figure*}

\begin{figure*}[h]
    \centering
    \includegraphics[width=\linewidth]{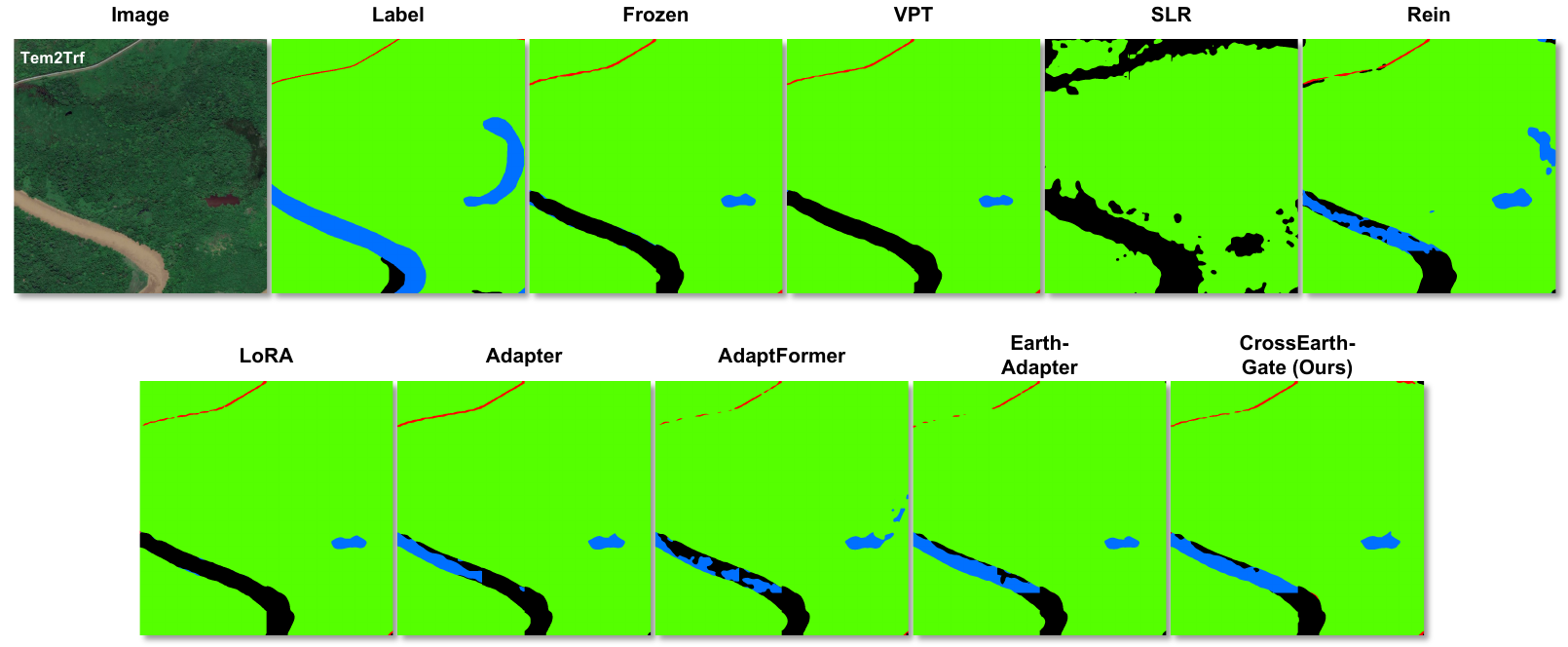}
    \caption{Complete visualizations of predicted segmentation maps of PEFT methods. 
    These samples are collected from the domain generalization benchmarks of Tem2Trf on the CASID \cite{liu2023large} dataset, where {\color{red} red} is the road class, {\color{yellow} yellow} is the building class, {\color{blue} blue} is the water class, {\color{green} green} is the forest class, and black is the background class.
    }
    \label{fig:tem2trf}
\end{figure*}

\begin{figure*}[t]
    \centering
    \includegraphics[width=\linewidth]{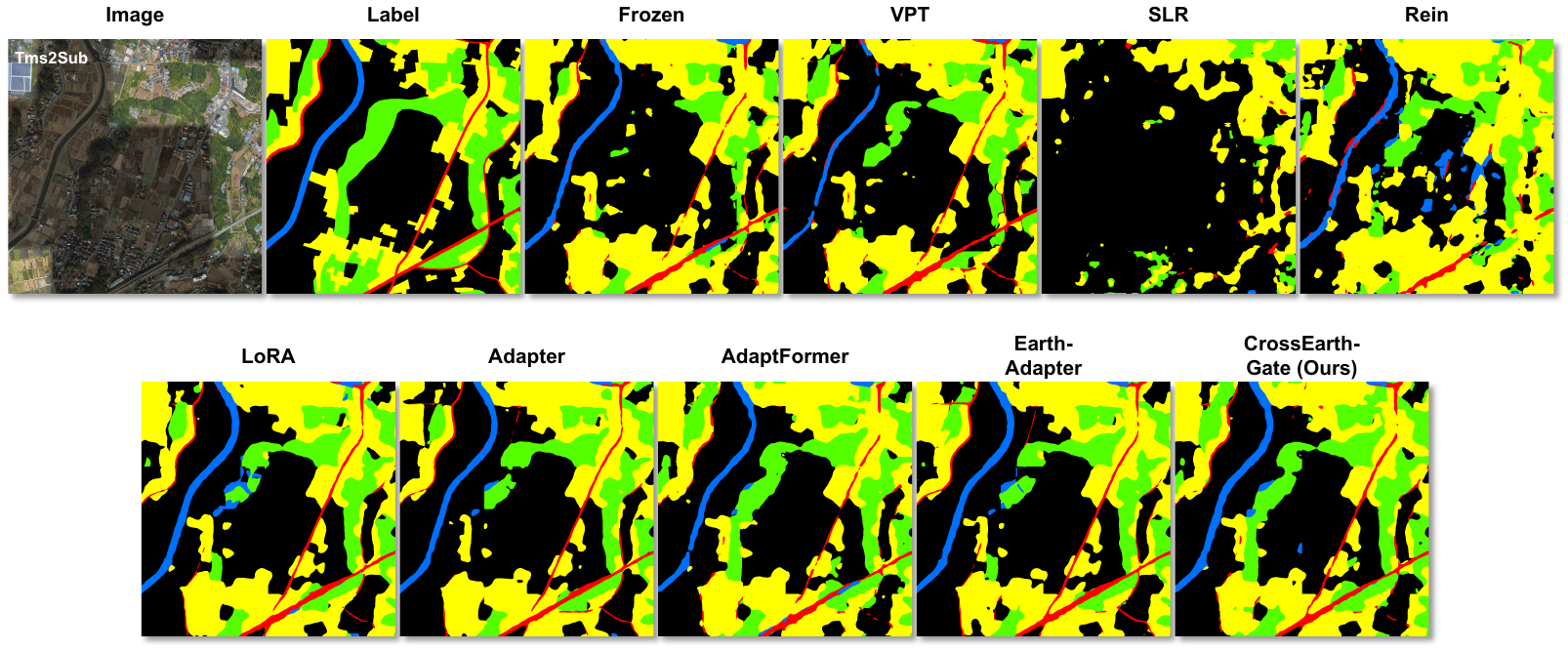}
    \caption{Complete visualizations of predicted segmentation maps of PEFT methods. 
    These samples are collected from the domain generalization benchmarks of Tms2Sub on the CASID \cite{liu2023large} dataset, where {\color{red} red} is the road class, {\color{yellow} yellow} is the building class, {\color{blue} blue} is the water class, {\color{green} green} is the forest class, and black is the background class.
    }
    \label{fig:tms2sub}
\end{figure*}

\begin{figure*}[t]
    \centering
    \includegraphics[width=\linewidth]{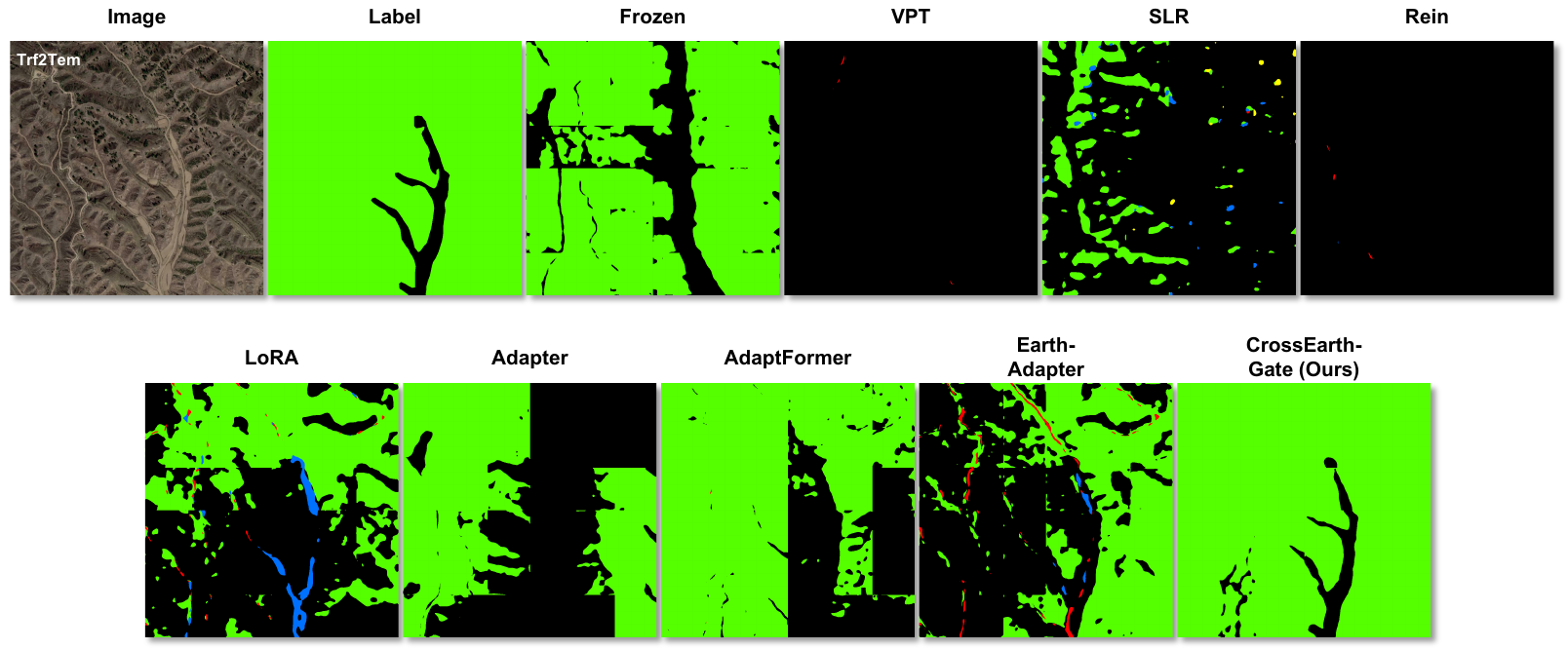}
    \caption{Complete visualizations of predicted segmentation maps of PEFT methods. 
    These samples are collected from the domain generalization benchmarks of Trf2Tem on the CASID \cite{liu2023large} dataset, where {\color{red} red} is the road class, {\color{yellow} yellow} is the building class, {\color{blue} blue} is the water class, {\color{green} green} is the forest class, and black is the background class.
    }
    \label{fig:trf2tem}
\end{figure*}

\begin{figure*}[t]
    \centering
    \includegraphics[width=\linewidth]{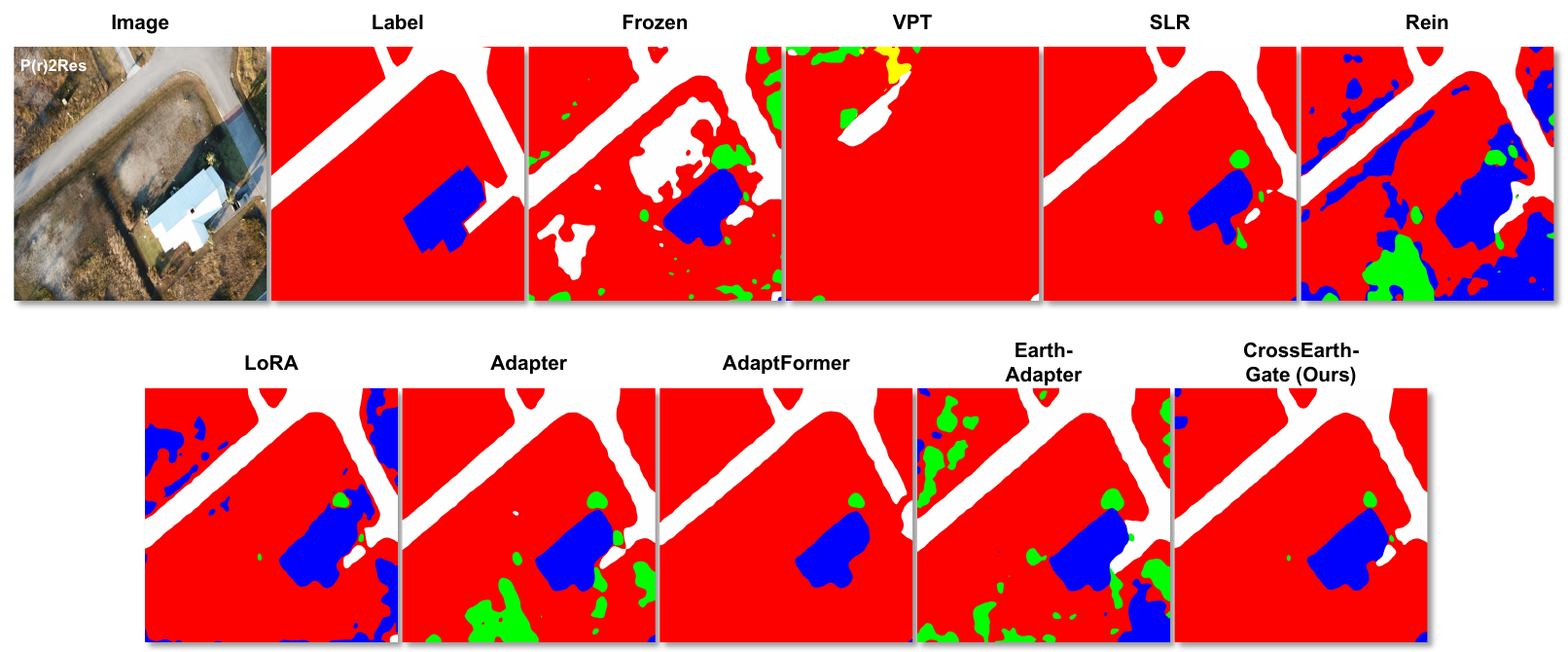}
    \caption{Complete visualizations of predicted segmentation maps of PEFT methods. 
    These samples are collected from the domain generalization benchmarks of P(r)2Res on the the RescueNet \cite{rahnemoonfar2023rescuenet} dataset, where white is the impervious surface class, {\color{red} red} is the clutter class, {\color{blue} blue} is the building class, {\color{green} green} is the vegetation class, and {\color{yellow} yellow} is the car class.
    }
    \label{fig:pr}
\end{figure*}

\begin{figure*}[t]
    \centering
    \includegraphics[width=\linewidth]{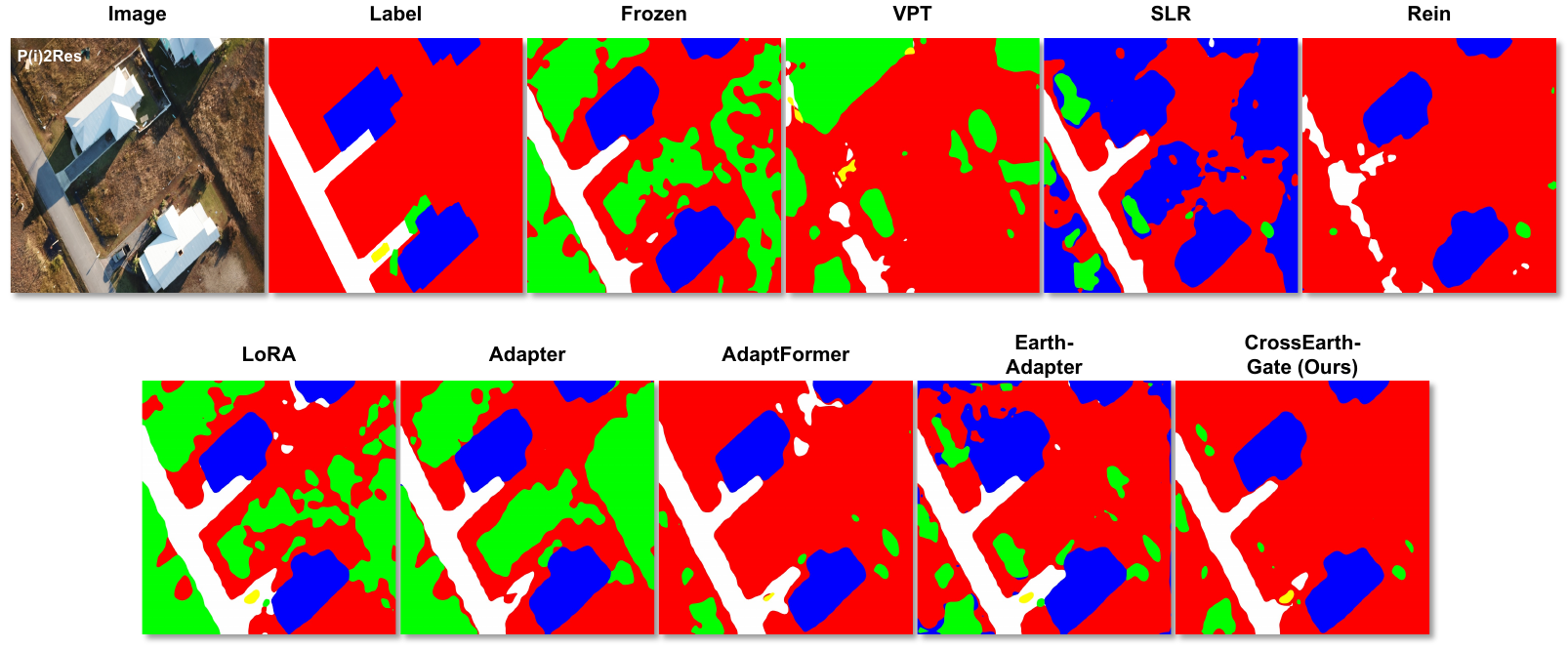}
    \caption{Complete visualizations of predicted segmentation maps of PEFT methods. 
    These samples are collected from the domain generalization benchmarks of P(i)2Res on the the RescueNet \cite{rahnemoonfar2023rescuenet} dataset, where white is the impervious surface class, {\color{red} red} is the clutter class, {\color{blue} blue} is the building class, {\color{green} green} is the vegetation class, and {\color{yellow} yellow} is the car class.
    }
    \label{fig:pi}
\end{figure*}

\begin{figure*}[t]
    \centering
    \includegraphics[width=\linewidth]{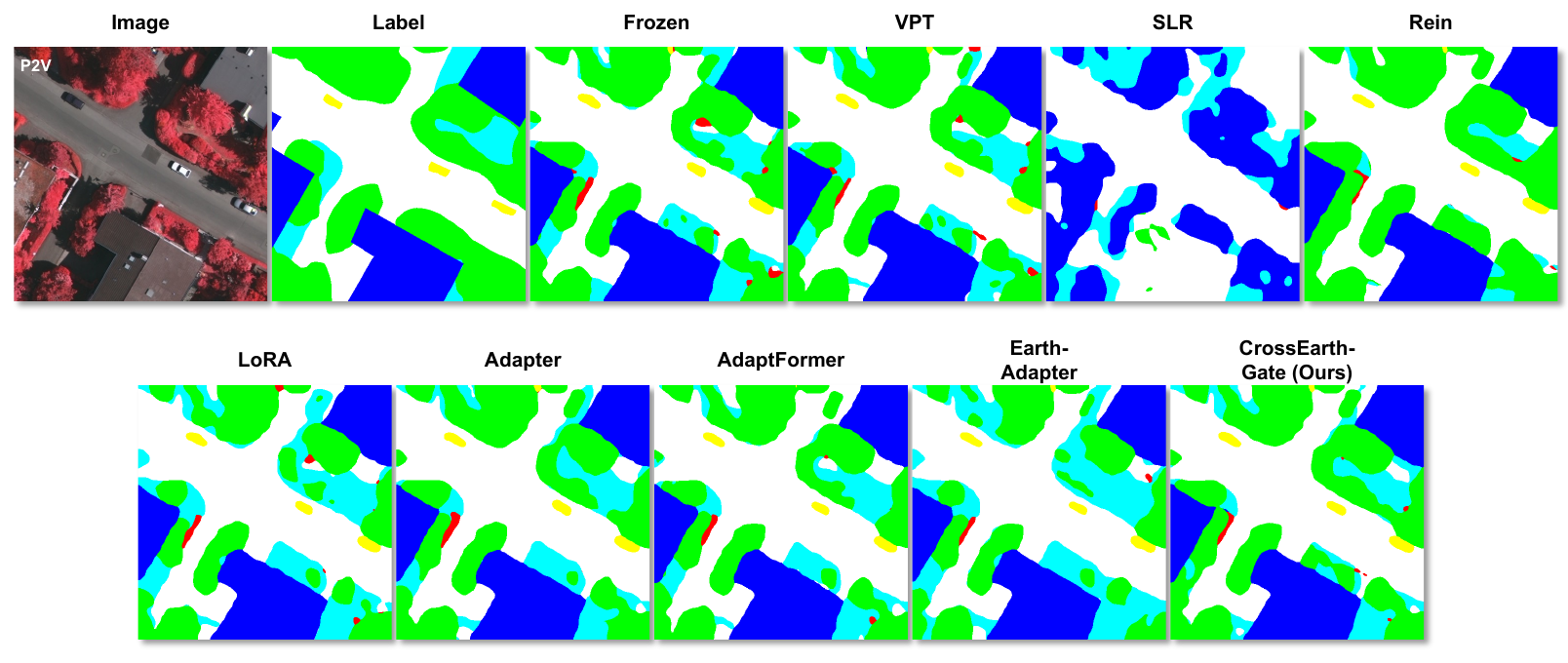}
    \caption{Complete visualizations of predicted segmentation maps of PEFT methods. 
    These samples are collected from the domain generalization benchmarks of P2V on the the Vaihingen dataset, where white is the impervious surface class, {\color{red} red} is the clutter class, {\color{blue} blue} is the building class, \textcolor{mycyan}{cyan} is the low vegetation, {\color{green} green} is the tree class, and {\color{yellow} yellow} is the car class.
    }
    \label{fig:p2v}
\end{figure*}

\begin{figure*}[t]
    \centering
    \includegraphics[width=\linewidth]{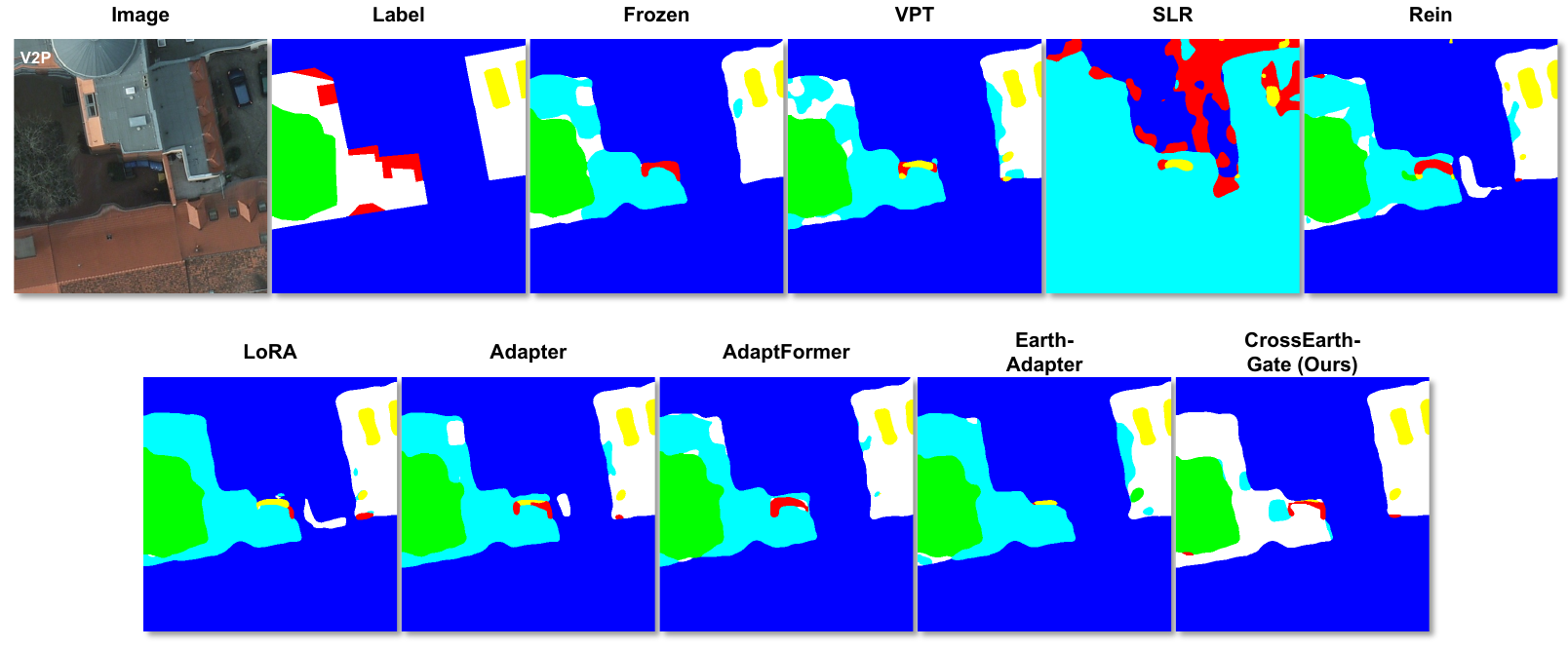}
    \caption{Complete visualizations of predicted segmentation maps of PEFT methods. 
    These samples are collected from the domain generalization benchmarks of P2V on the the Potsdam dataset, where white is the impervious surface class, {\color{red} red} is the clutter class, {\color{blue} blue} is the building class, \textcolor{mycyan}{cyan} is the low vegetation, {\color{green} green} is the tree class, and {\color{yellow} yellow} is the car class.
    }
    \label{fig:v2p}
\end{figure*}

\begin{figure*}[t]
    \centering
    \includegraphics[width=\linewidth]{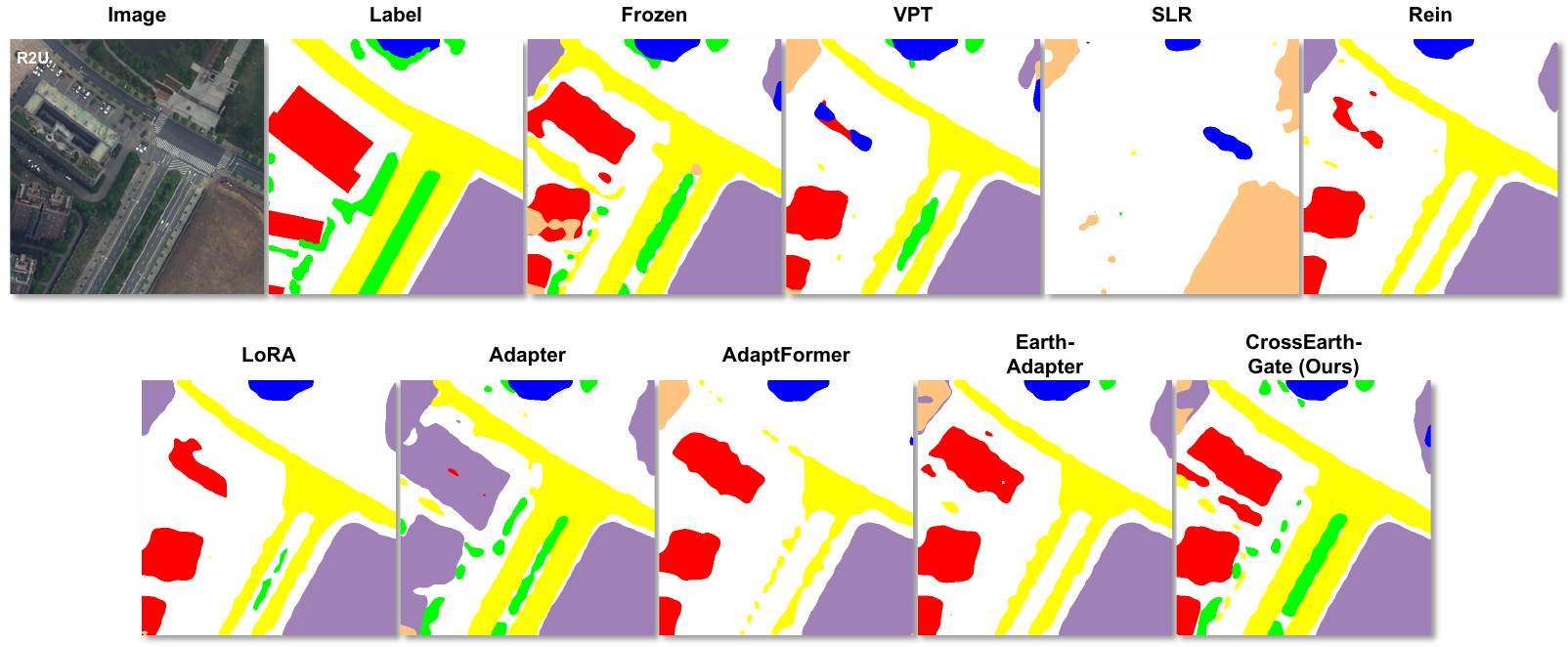
    }
    \caption{Complete visualizations of predicted segmentation maps of PEFT methods. 
    These samples are collected from the domain generalization benchmarks of R2U on the the LoveDA \cite{wang2021loveda} dataset, where white is the background class, {\color{red} red} is the building class, {\color{yellow} yellow} is the road class, {\color{blue} blue} is the water class, {\color{purple} purple} is the barren class, {\color{green} green} is the forest class, and {\color{mybrown} brown} is the agriculture class.
    }
    \label{fig:r2u}
\end{figure*}

\begin{figure*}[t]
    \centering
    \includegraphics[width=\linewidth]{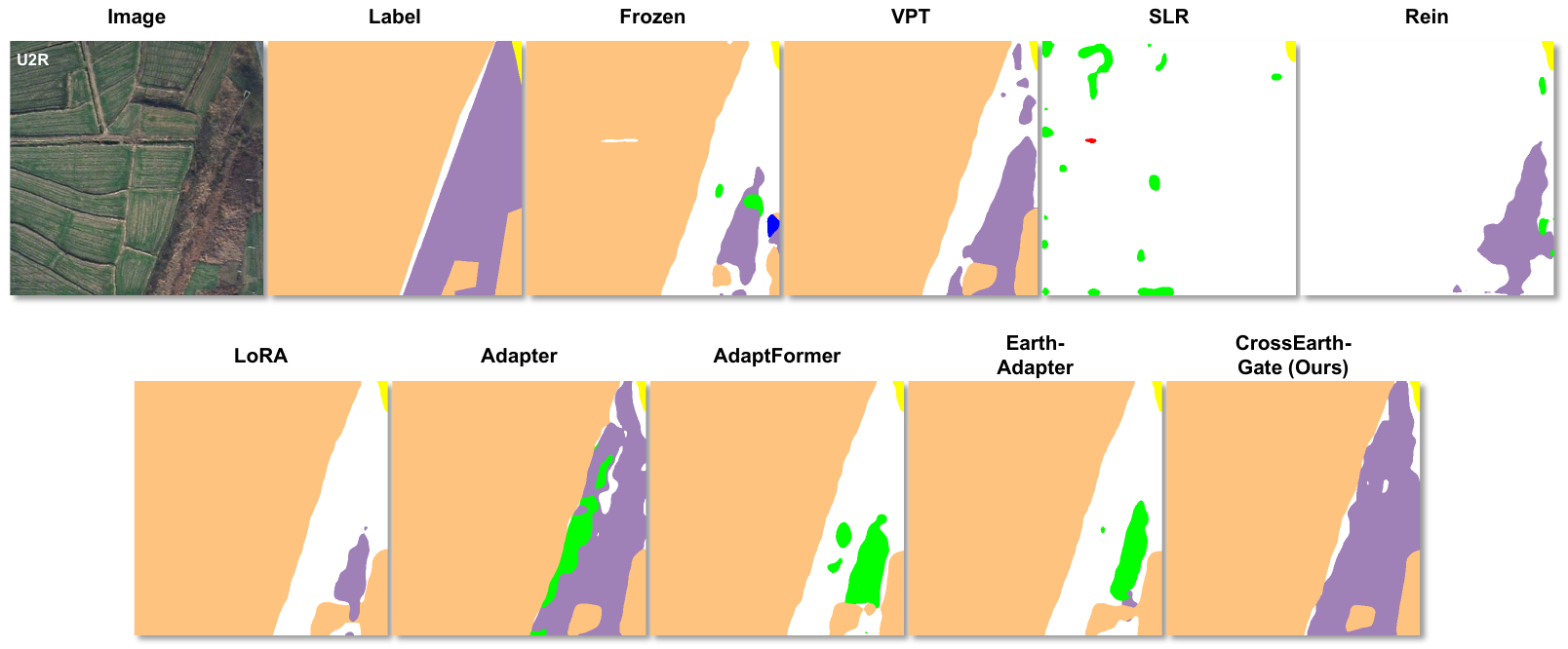
    }
    \caption{Complete visualizations of predicted segmentation maps of PEFT methods. 
    These samples are collected from the domain generalization benchmarks of U2R on the the LoveDA \cite{wang2021loveda} dataset, where white is the background class, {\color{red} red} is the building class, {\color{yellow} yellow} is the road class, {\color{blue} blue} is the water class, {\color{purple} purple} is the barren class, {\color{green} green} is the forest class, and {\color{mybrown} brown} is the agriculture class.
    }
    \label{fig:u2r}
\end{figure*}

The CASID benchmarks (Fig. \ref{fig:sub2tms} - \ref{fig:trf2tem}) highlight the acute limitations of single-pathway adaptation when facing simultaneous spectral and geometric shifts.
For example, in the Sub2Tms scenarios, the domain shift involves significant changes in vegetation appearance and water surface texture (frequency artifacts). 
Methods like AdaptFormer and LoRA, which specialize in semantic features or spatial, struggle to distinguish between large water bodies and forests when high-frequency spectral noise is present.
This results in substantial ``hallucinations'' of land over water.
The Tem2Trf benchmarks introduce complex geometric shifts in road networks. 
As seen in Fig. \ref{fig:tem2trf}, methods lacking strong spatial reasoning, such as AdaptFormer and Earth-Adapter, frequently produce fragmented road predictions, failing to maintain connectivity.
As shown in Fig. \ref{fig:pr} and \ref{fig:pi}, the transition from standard aerial imagery (Potsdam) to post-disaster scenes (RescueNet) presents a severe semantic challenge.
While LoRA may preserve object boundaries, it misinterprets the texture of disaster debris, misclassifying ``clutter'' (red) as ``vegetation'' (green) in the P(i)2Res benchmark (Fig. \ref{fig:pi}). 
This represents a critical semantic error where the spatial adaptation is insufficient to correct the conceptual drift.
The results for Domain Adaptation (DA) tasks (P2V, V2P, R2U, U2R), visualized in Fig. \ref{fig:p2v} - \ref{fig:u2r}, further validate our approach in scenarios with access to unlabeled target data.
In the transition between urban and rural domains (e.g., U2R in Fig. \ref{fig:u2r}), existing methods struggle to reconcile the scale differences between dense urban clusters and sparse rural features. For instance, AdaptFormer struggles to maintain the structural integrity of roads, while Earth-Adapter fails to resolve the boundaries of dense, small-scale forest regions. These failures indicate that a fixed adaptation strategy cannot dynamically prioritize the spatial refinement needed for scale variations or the semantic tuning needed for layout changes.

In contrast to these baselines, CrossEarth-Gate effectively navigates these trade-offs. 
By leveraging Fisher Information to quantify the importance of each module dynamically, our engine activates the appropriate pathway for the specific shift at hand.
Consequently, our method consistently produces segmentation maps that are closer to the ground-truth labels across all scenarios, demonstrating a robust and superior generalization capability.

\clearpage

\clearpage

\clearpage

\clearpage

\section{Additional Experimental Details}
\label{details}
In this section, we provide a comprehensive breakdown of the datasets and experiments utilized to benchmark CrossEarth-Gate. To ensure a rigorous evaluation of the multifaceted domain shifts inherent to remote sensing, we select datasets representing distinct challenges: climate variation, spectral band discrepancies, urban-rural transitions, and post-disaster alterations. The detailed statistics and configurations for each benchmark, including image resolution, sample sizes, and semantic categories, are summarized in Table \ref{benchmark}.
All models are fine-tuned using the AdamW \cite{loshchilov2017decoupled} optimizer, employing a weight decay of 0.05 and betas as 0.9 and 0.999.
We employ the data augmentation techniques following \cite{hu2025earth,wei2024stronger}.
The parameter fluctuation in the all table represents the range of the active trainable parameter set.

\subsection{Dataset Introduction}

\input{sec/tab/tab6}

\subsubsection{CASID}
The Climate-Aware Satellite Image Dataset (CASID)  \cite{liu2023large} represents a pioneering advancement in the field of RS. 
It is recognized as the first dataset explicitly engineered to mitigate domain shift challenges stemming from climatic diversity. 
By systematically capturing distinct environmental variances, CASID enables the development of models that are robust across shifting geographic and meteorological domains.
The dataset is stratified across four primary climate zones: Subtropical Monsoon (Sub), Temperate Monsoon (Tem), Tropical Monsoon (Tms), and Tropical Rainforest (Trf).
To facilitate precise land-cover segmentation, CASID provides pixel-wise annotations across five semantic categories. These classes capture the essential morphological and spectral features required for earth observation tasks, including Background, Building, Forest, Road, and Water.
We divide the dataset into training and validation sets and crop images to 1024 $\times$ 1024 resolutions, following the data processing protocols of \cite{gong2024crossearth}.
For the CASID DG experiment, we fine-tune the model for 30k iterations with a batch size of 1.
\subsubsection{ISPRS Potsdam and Vaihingen}
The ISPRS Potsdam and Vaihingen datasets are widely recognized as foundational benchmarks in the field of RS semantic segmentation. 
Curated by the International Society for Photogrammetry and Remote Sensing (ISPRS), these datasets consist of High Spatial Resolution (HSR) aerial imagery captured over their respective namesake cities in Germany. 
Due to their distinct environmental and spectral characteristics, these datasets have become the standard for evaluating Domain Adaptation (DA) techniques \cite{zhang2023pseudo,liang2023multilevel,ma2023unsupervised,zhao2023domain,wang2023fine,hu2025earth}. 
They present a robust challenge for algorithms attempting to bridge the domain gap between differing urban landscapes and sensor configurations.
The annotations cover six common land-cover categories: Impervious surfaces, Building, Low vegetation, Tree, Car, and Clutter.
We utilize the RGB channels of Potsdam as the source domain and the IR-R-G channels of Vaihingen as the target domain (P2V), and vice-versa (V2P) with a cropped size of 512 $\times$ 512, following \cite{hu2025earth}.
For the DA experiment, we fine-tune the model for 20k iterations with a batch size of 2.
\subsubsection{LoveDA}
The LoveDA (Land-cOVEr Domain Adaptive semantic segmentation) \cite{wang2021loveda} dataset is a large-scale HSR benchmark designed to advance both semantic segmentation and transferable learning. 
Unlike traditional datasets that focus solely on semantic representation, LoveDA explicitly addresses DA challenges by capturing the diverse stylistic and structural differences between developed and undeveloped geographic areas.
The dataset comprises 5,987 HSR images collected from three different Chinese cities (Nanjing, Changzhou, and Wuhan). These images cover 536.15 km$^2$ at a spatial resolution of 0.3 meters.
Specifically, LoveDA is uniquely structured around two distinct domains, which introduce significant challenges regarding multi-scale objects and inconsistent class distributions: Urban domain, characterized by high population density, neatly arranged buildings of various shapes, and wide roads, and Rural domain, characterized by natural elements, disordered building layouts, small-scale agricultural zones, and narrow roads.
We follow the protocol of \cite{hu2025earth} to establish two DA settings: Urban-to-Rural (U2R) and Rural-to-Urban (R2U). 
The evaluation is performed across seven semantic categories: Background, Building, Road, Water, Barren, Forest, and Agriculture.
For the DA experiment, we fine-tune the model for 20k iterations with a batch size of 2.
\subsubsection{RescueNet}
RescueNet \cite{rahnemoonfar2023rescuenet} stands as a cutting-edge, high-resolution benchmark designed to advance the capabilities of Unmanned Aerial Vehicles (UAVs) in post-disaster scenarios. 
Unlike satellite-based datasets that often suffer from low resolution, RescueNet consists of low-altitude aerial imagery captured specifically after Hurricane Michael.
The primary objective of RescueNet is to overcome the limitations of existing datasets by providing: (1) holistic scene understanding, which moves beyond simple building detection to provide pixel-level annotations for the entire scene, including infrastructure and natural elements; and (2) granular damage assessment, which introduces detailed severity classifications for buildings and roads, enabling rescue teams to prioritize efforts based on precise damage levels.
RescueNet contains 11 categories, including Background, Water, Building-No-Damage, Building-Medium-Damage, Building-Major-Damage, Building-Total-Destruction, Vehicle, Road-Clear, Road-Blocked, Tree, and Pool.
In our experiments, we utilize RescueNet solely as an unseen target domain, using the RGB (P(r)2Res) and IR-R-G (P(i)2Res)channels Potsdam dataset as the source, following the protocol from \cite{gong2024crossearth}. 
This setup assesses the model's ability to maintain semantic integrity when transferring from organized urban environments to chaotic, post-disaster scenes. The evaluation focuses on five shared classes: Impervious surfaces, Building, Tree, Car, and Clutter.
For the DG experiment, we fine-tune the model for 30k iterations with a batch size of 1.
\subsection{PEFT Baseline Implementation}
\input{sec/tab/tab7}

We implement eight PEFT baseline methods to rigorously benchmark the performance of our proposed CrossEarth-Gate. 
The default hyperparameters for these methods, including rank configurations, prompt lengths, and scaling factors, are detailed in Table \ref{baseline}. Below, we briefly describe the implementation specifics for each baseline.
\subsubsection{VPT}
Visual Prompt Tuning (VPT) \cite{jia2022visual} is utilized to adapt the frozen foundation model by injecting a sequence of learnable parameters, known as soft prompts, directly into the input sequence. 
Instead of fine-tuning the backbone weights, VPT prepends these trainable tokens to the sequence of image patch embeddings. 
These prompts interact with the image features via the Transformer's self-attention mechanisms, effectively encoding task-specific context to steer the model's representation. 
We implement the ``Deep'' version of VPT, where learnable prompt tokens are inserted into the sequence at every Transformer layer. 
As indicated in Table \ref{baseline}, we set the prompt length to 128 tokens and disable dropout to maintain signal integrity during the adaptation of geospatial features.
\subsubsection{SLR}
We implement the Scaled Low-Rank (SLR) adapter method as part of our PEFT baseline, which introduces a small number of parameters to enhance pre-trained transformer models for domain adaptation, especially in remote sensing applications. 
The SLR method focuses on efficiently adapting foundation models without retraining all parameters. 
It achieves this by introducing low-rank matrices and learnable scaling vectors into the existing transformer layers, specifically targeting the linear transformations in the Multi-head Self-Attention (MSA) and Multi-Layer Perceptron (MLP) components. 
The key advantage of SLR is its ability to scale the model’s capacity with minimal computational overhead by adjusting only the scaling parameters and the low-rank matrices. 
We set the bottleneck dimension to 16, focusing on parameter efficiency while allowing sufficient capacity for modality alignment.

\subsubsection{Rein}
Rein \cite{wei2024stronger} is a robust fine-tuning method specifically designed to harness Vision Foundation Models for domain generalization tasks. 
Unlike standard adapters that insert modules within the Transformer blocks, Rein refines the output feature maps of each frozen backbone layer. 
It introduces a set of learnable tokens that interact with the image features through a dot-product similarity mechanism to generate instance-level feature refinements. 
To maintain parameter efficiency, the method employs a low-rank decomposition for the token sequences ($T = A \times B$) and shares the MLP weights across all layers. 
Following the default settings of the official code, we configure the token length to 100 and the low-rank dimension to 16.  
Additionally, we enable the explicit linkage between these learnable tokens and the object queries in the decoder to further enhance instance discrimination.


\subsubsection{LoRA}
Low-Rank Adaptation (LoRA) \cite{hu2022lora} freezes the pre-trained model weights and injects trainable rank decomposition matrices into the self-attention layers. 
We specifically target the Query and Value projection matrices in the MSA blocks. 
We set the rank $r$ to 64 and the scaling factor $\alpha$ to 1, ensuring that the low-rank updates can sufficiently capture spatial dependencies without introducing excessive parameters.
\subsubsection{Adapter}
The Adapter \cite{houlsby2019parameter} method, originally proposed for efficient natural language processing transfer learning, introduces small bottleneck modules between the pre-trained layers of a Transformer model. 
Unlike full fine-tuning, which updates all parameters, Adapters freeze the pre-trained weights and only train the newly added parameters, typically comprising a very small percentage of the original model size. 
In the standard architecture, two adapter modules are inserted per Transformer layer: one after the multi-head attention projection and another after the feed-forward network. 
Each adapter consists of a down-projection to a low-dimensional bottleneck, a non-linear activation, and an up-projection back to the original dimension, along with a skip-connection to facilitate identity initialization. 
We set the bottleneck dimension to 64 and apply a dropout rate of 0.1.

\subsubsection{AdaptFormer}
AdaptFormer \cite{chen2022adaptformer} is a method designed to efficiently adapt Vision Transformers \cite{dosovitskiy2020image} (ViTs) to downstream tasks by replacing the original MLP block with a modified ``AdaptMLP'' module. 
This module introduces a lightweight bottleneck structure in parallel to the frozen MLP layers. 
Specifically, the AdaptMLP consists of two branches: the original frozen MLP and a trainable branch comprising a down-projection layer, a non-linear activation (ReLU), and an up-projection layer.  
The trainable branch processes the input features and scales the output by a learnable factor $s$ before adding it to the frozen MLP output via a residual connection. 
This design allows the model to learn task-specific features with a minimal number of additional parameters. 
In our experiments, we set the bottleneck dimension to 64 and initialize the scaling factor $s$ to 0.1.

\subsubsection{Earth-Adapter}
Earth-Adapter \cite{hu2025earth} is designed to mitigate the impact of high-frequency artifacts prevalent in RS imagery. 
It employs a Mixture of Frequency Adaptation mechanisms, which integrates a Mixture of Adapters (MoA) with the Discrete Fourier Transform (DFT). 
The module operates in parallel to the frozen backbone layers. 
Specifically, the input features are processed by three expert branches: a standard Spatial Adapter, a Low-Frequency (LF) Adapter, and a High-Frequency (HF) Adapter.  
The frequency adapters utilize DFT to decompose features based on a learnable cutoff frequency $\rho$, separating artifacts (typically high-frequency) from global semantic structures (low-frequency). 
A dynamic router then assigns weights to these experts to fuse the features.

To ensure optimal performance, we strictly follow the hyperparameter configurations derived from the extensive ablation studies reported in the original paper for the Domain Adaptation (DA) benchmarks. 
For Potsdam $\to$ Vaihingen (P2V), we set the bottleneck dimension to 64, cutoff frequency to 0.3, and apply frequency adapters to shallow layers [0-2]. 
For Vaihingen $\to$ Potsdam (V2P), we use dimension 32, cutoff 0.2, and deep layers [21-23]. 
For Rural $\to$ Urban (R2U), we use dimension 16, cutoff 0.3, and layers [21-23]. 
For Urban $\to$ Rural (U2R), we use dimension 32, cutoff 0.2, and layers [18-23]. 
For all Domain Generalization (DG) experiments, we utilize the default robust setting with a bottleneck dimension of 64, a cutoff frequency of 0.3, and layers [0-23]. 
We follow the official code to implement the Earth-Adapter, with the initial scaling factor set to 0.1 across all experiments.

\subsection{Proposed Methods}
The pseudo-code for CrossEarth-Gate is described in Algorithm \ref{alg:crossearth_gate}. 
Furthermore, we provide the default hyperparameters for our proposed CrossEarth-Gate in Table \ref{gate}. 
The implementation is based on the MMEngine framework. 
CrossEarth-Gate introduces a dynamic selection mechanism that activates a specific subset of modules from the RS Module Toolbox based on Fisher Information.
The importance scores are calculated using the squared gradients accumulated over a defined number of steps, normalized relatively across module types to ensure balanced selection.

\input{sec/tab/tab8}

\begin{algorithm}[h]
\small
\caption{Pseudo-code for CrossEarth-Gate}
\label{alg:crossearth_gate}
\textbf{Input:} Pre-trained foundation model $\alpha$, RS module toolbox $\zeta$ (Spatial, Semantic, Frequency), training dataset $D$. \\
\textbf{Parameters:} Evaluation samples $M$, update interval $N$, maximum active modules $k$.
\begin{algorithmic}[1]
\STATE \textbf{Initialize:} Insert all modules $\zeta$ into every block of $\alpha$ and set $Iter \leftarrow 0$.
\STATE \textbf{Set State:} $EvaluatePhase \leftarrow \text{True}$.
\WHILE{training not converged}
    \STATE Zero gradients.
    \STATE Sample batch $(\mathbf{X}, \mathbf{Y})$ from $D$.
    \STATE Compute task loss $\mathcal{L}$ and backpropagate to obtain gradients.
    
    \IF{$EvaluatePhase$ is True}
        \FOR{each module parameter $\zeta_{i}^{z}$ in layer $i$ of type $z$}
            \STATE Accumulate Fisher Info: $\hat{F}_{\zeta_i^z} \mathrel{+}= (\nabla_{\zeta_i^z} \mathcal{L})^2$
        \ENDFOR
        
        \IF{evaluated $M$ samples}
            \FOR{each module type $z$ and layer $i$}
                \STATE Aggregate score: $\hat{S}_i^z \leftarrow \sum_{\zeta_i^z} \hat{F}_{\zeta_i^z}$
            \ENDFOR
            \FOR{each module type $z$}
                \STATE Relative score: $S_i^z \leftarrow \hat{S}_i^z / \sum_{j} \hat{S}_j^z$
            \ENDFOR
            \STATE $\text{Activate Top-}k \text{ modules with the highest } S_i^z$ and freeze all other modules.
            \STATE $EvaluatePhase \leftarrow \text{False}$, Reset $\hat{F} \leftarrow 0$.
        \ENDIF
    \ELSE
        \STATE Update weights of active modules.
        
        \IF{$Iter \mod N == 0$}
            \STATE Activate gradients for all modules in toolbox.
            \STATE $EvaluatePhase \leftarrow \text{True}$.
        \ENDIF
        \STATE $Iter \leftarrow Iter + 1$
    \ENDIF
\ENDWHILE
\end{algorithmic}
\end{algorithm}

%% file: sec/tab/tab5.tex
\begin{table*}[t]
\caption{
Complete ablation studies on CASID benchmarks across 12 DG experiments. 
We demonstrate the generalizability of CrossEarth-Gate across different backbones and compare the performance and trainable parameters of CrossEarth-Gate against versions with key components removed.
Sub: Subtropical Monsoon. Tem: Temperate Monsoon. Tms: Tropical Monsoon. Trf: Tropical Rainforest.}
\centering
\resizebox{\textwidth}{!}{
\begin{tabular}{ccc|cccc|cccc}
\toprule
Backbone & Method &  Params (M) & Sub2Tem & Sub2Tms & Sub2Trf &Average & Tem2Sub & Tem2Tms & Tem2Trf & Average\\ 
\midrule
 
\multirow{3}{*}{
{\begin{tabular}[c]{@{}c@{}} SatMAE \\ (Large) \cite{cong2022satmae} \end{tabular}}
} 
& Frozen &0.0 &22.4  &24.6  &26.1 &24.4 &22.3 &18.9 &\textbf{23.3} &21.5\\
& Full-Tuning &304.2 &\textbf{26.7} &42.2 &37.3 &35.4 &11.1 &16.1 &17.4 &14.9\\
&\cellcolor{lightgreen}\textbf{CrossEarth-Gate}  &\cellcolor{lightgreen}2.9-4.0  &\cellcolor{lightgreen}21.8  &\cellcolor{lightgreen}\textbf{46.8} &\cellcolor{lightgreen}\textbf{42.9} 
&\cellcolor{lightgreen}\textbf{37.1} &\cellcolor{lightgreen}\textbf{29.2} &\cellcolor{lightgreen}\textbf{21.6} &\cellcolor{lightgreen}14.6 &\cellcolor{lightgreen}\textbf{21.8} \\
\midrule

\multirow{3}{*}{
{\begin{tabular}[c]{@{}c@{}} Scale-MAE \\ (Large) \cite{reed2023scale} \end{tabular}}
} 
& Frozen &0.0 &38.6  &58.9  &54.6 &50.7 &55.6 &50.9 &38.9 &48.5\\
&  Full-Tuning &304.2  &38.7 &64.0 &55.8 &52.8 &62.7 &59.2 &\textbf{51.0} &57.6\\
&\cellcolor{lightgreen} \textbf{CrossEarth-Gate}  &\cellcolor{lightgreen} 2.9-3.6  &\cellcolor{lightgreen}\textbf{45.1} &\cellcolor{lightgreen}\textbf{66.3} &\cellcolor{lightgreen}\textbf{57.4} &\cellcolor{lightgreen}\textbf{56.3} &\cellcolor{lightgreen}\textbf{64.6} &\cellcolor{lightgreen}\textbf{61.0} &\cellcolor{lightgreen}48.7 &\cellcolor{lightgreen}\textbf{58.1}\\
\midrule

\multirow{3}{*}{
{\begin{tabular}[c]{@{}c@{}} SAM \\(Huge) \cite{kirillov2023segment}\end{tabular}}
} 
& Frozen                    &0.0   &39.2  &\textbf{64.0}  &56.7 &53.3 &63.5 &61.9 &47.1 &57.5\\
&  Full-Tuning              &631.2 &\textbf{43.3}  &60.6  &60.2 &54.7 &63.1 &60.2 &51.2 &58.2\\
&\cellcolor{lightgreen}\textbf{CrossEarth-Gate} &\cellcolor{lightgreen}4.0-4.9       &\cellcolor{lightgreen}42.9 &\cellcolor{lightgreen}63.7 &\cellcolor{lightgreen}\textbf{61.7} &\cellcolor{lightgreen}\textbf{56.1} &\cellcolor{lightgreen}\textbf{67.1} &\cellcolor{lightgreen}\textbf{65.4} &\cellcolor{lightgreen}\textbf{53.3} &\cellcolor{lightgreen}\textbf{61.9} \\
\midrule

\multirow{3}{*}{
{\begin{tabular}[c]{@{}c@{}} DINOv2\\(Small)  \cite{oquab2023dinov2}\end{tabular}}
} 
 & Frozen      &0.0   &43.0  &50.9 &54.7  &49.5  &64.9 &58.8 &55.1 &59.6\\
&  Full-Tuning &22.1  &33.9  &\textbf{60.2} &54.2  &49.4   &56.5 &49.6 &41.9 &49.3\\

 &\cellcolor{lightgreen}\textbf{CrossEarth-Gate}  &\cellcolor{lightgreen}0.7  &\cellcolor{lightgreen}\textbf{46.2} &\cellcolor{lightgreen}58.4 &\cellcolor{lightgreen}\textbf{59.1} &\cellcolor{lightgreen}\textbf{54.6} &\cellcolor{lightgreen}\textbf{65.1} &\cellcolor{lightgreen}\textbf{63.9} &\cellcolor{lightgreen}\textbf{57.1} &\cellcolor{lightgreen}\textbf{62.0} \\
\midrule

\multirow{3}{*}{
{\begin{tabular}[c]{@{}c@{}} DINOv2 \\(Base) \cite{oquab2023dinov2}\end{tabular}}
} 
& Frozen                    &0.0        &39.3 &57.2 &57.4 &51.3 &\textbf{67.2} &63.9 &56.4 &62.5 \\
& Full-Tuning               &86.6       &36.3 &59.3 &54.8 &50.2 &58.6 &54.9 &47.5 &53.7 \\
& \cellcolor{lightgreen}\textbf{CrossEarth-Gate}  &\cellcolor{lightgreen}1.4-1.8    & \cellcolor{lightgreen}\textbf{45.3} &\cellcolor{lightgreen}\textbf{62.5} &\cellcolor{lightgreen}\textbf{58.9} &\cellcolor{lightgreen}\textbf{55.6} &\cellcolor{lightgreen}\textbf{67.2} &\cellcolor{lightgreen}\textbf{65.7} &\cellcolor{lightgreen}\textbf{57.9} &\cellcolor{lightgreen}\textbf{63.6}\\
\midrule

& Frozen & 0.0 &43.1 &63.4 &58.8 &55.2 &66.9 &64.8 &59.0 &63.6  \\
& Full-Tuning &304.2 &38.6 &62.3 &58.3 &53.0 &58.7 &55.8 &43.8 &52.8 \\
\rowcolor{lightgreen} \cellcolor{white}
& \textbf{CrossEarth-Gate} &3.0-4.4   &50.1 &\textbf{66.6} &\textbf{65.2} &\textbf{60.6} &68.0 &\textbf{67.0} &\textbf{60.3} &\textbf{65.1}  \\
&w/o Spatial   &2.7-3.2  &47.9 &60.3 &61.9 &56.7 &68.4 &66.8 &59.5 &64.8\\
&w/o Semantic  &4.2-4.5  &51.3 &\textbf{66.6} &60.2 &59.4 &\textbf{68.5} &66.2 &57.9 &64.2\\
&w/o Frequency &3.2-4.2  &\textbf{51.4} &57.6 &63.4 &57.5 &67.5 &64.0 &59.9 &63.8\\
\multirow{-7}{*}{
{\begin{tabular}[c]{@{}c@{}} DINOv2\\(Large) \cite{oquab2023dinov2}\end{tabular}}} 
&w/o Selection &14.4  &48.0 &61.8 &63.5 &57.8 &67.7 &63.6 &59.2 &63.5\\

\midrule

Backbone & Method &  Params (M) &Tms2Sub  &Tms2Tem &Tms2Trf &Average & Trf2Sub & Trf2Tem & Trf2Tms & Average\\ 
 \midrule

\multirow{3}{*}{
{\begin{tabular}[c]{@{}c@{}} SatMAE \\ (Large) \cite{cong2022satmae} \end{tabular}}
} 
& Frozen &0.0 &28.8 &18.5  &27.0 &24.8 &33.1 &\textbf{22.0} &23.7 &26.3\\
&  Full-Tuning &304.2  &42.0 &\textbf{24.6} &39.1 &35.2 &37.7 &20.0 &\textbf{31.8} &29.8\\
&\cellcolor{lightgreen}\textbf{CrossEarth-Gate}  &\cellcolor{lightgreen}2.9-4.0  &\cellcolor{lightgreen}\textbf{45.2} &\cellcolor{lightgreen}24.1 &\cellcolor{lightgreen}\textbf{41.0} &\cellcolor{lightgreen}\textbf{36.8} &\cellcolor{lightgreen}\textbf{45.1} &\cellcolor{lightgreen}18.9 &\cellcolor{lightgreen}31.2 &\cellcolor{lightgreen}\textbf{31.8} \\
\midrule

\multirow{3}{*}{
{\begin{tabular}[c]{@{}c@{}} Scale-MAE \\ (Large) \cite{reed2023scale} \end{tabular}}
} 
& Frozen &0.0 &64.8  &31.5  &56.4 &50.9 &59.4 &34.5 &58.8 &50.9\\
&  Full-Tuning &304.2  &64.3 &33.0 &58.5 &51.9 &56.6 &\textbf{42.8} &58.6 &52.6\\
&\cellcolor{lightgreen} \textbf{CrossEarth-Gate}  &\cellcolor{lightgreen} 2.9-3.6  &\cellcolor{lightgreen}\textbf{67.3} &\cellcolor{lightgreen}\textbf{37.6} &\cellcolor{lightgreen}\textbf{59.6} &\cellcolor{lightgreen}\textbf{54.9} &\cellcolor{lightgreen}\textbf{62.4} &\cellcolor{lightgreen}36.3 &\cellcolor{lightgreen}\textbf{62.0} 
&\cellcolor{lightgreen}\textbf{53.8} \\
\midrule

\multirow{3}{*}{
{\begin{tabular}[c]{@{}c@{}} SAM \\(Huge) \cite{kirillov2023segment}\end{tabular}}
} 
& Frozen                    &0.0   &64.9  &37.7  &57.2 &53.2 &63.3 &34.2 &62.6 &53.4\\
&  Full-Tuning              &631.2 &64.6  &33.2  &57.4 &51.7 &63.9 &36.9 &62.9 &54.6\\
&\cellcolor{lightgreen}\textbf{CrossEarth-Gate} &\cellcolor{lightgreen}4.0-4.9       &\cellcolor{lightgreen}\textbf{65.8} &\cellcolor{lightgreen}\textbf{43.4} &\cellcolor{lightgreen}\textbf{59.9} &\cellcolor{lightgreen}\textbf{56.3} &\cellcolor{lightgreen}\textbf{67.2} &\cellcolor{lightgreen}\textbf{43.3} &\cellcolor{lightgreen}\textbf{67.3} &\cellcolor{lightgreen}\textbf{59.3} \\
\midrule

\multirow{3}{*}{
{\begin{tabular}[c]{@{}c@{}} DINOv2\\(Small)  \cite{oquab2023dinov2}\end{tabular}}
} 
 & Frozen      &0.0   &63.1  &\textbf{37.2} &55.5  &51.9  &63.5 &33.2 &61.5 &52.7\\
&  Full-Tuning &22.1  &60.5  &29.4  &56.4  &48.7 &58.9 &32.3 &59.5 &50.2\\

 &\cellcolor{lightgreen}\textbf{CrossEarth-Gate}  &\cellcolor{lightgreen}0.7  &\cellcolor{lightgreen}\textbf{64.6} &\cellcolor{lightgreen}35.0 &\cellcolor{lightgreen}\textbf{58.6} &\cellcolor{lightgreen}\textbf{52.7} &\cellcolor{lightgreen}\textbf{64.8} &\cellcolor{lightgreen}\textbf{38.1} &\cellcolor{lightgreen}\textbf{64.6} &\cellcolor{lightgreen}\textbf{55.8} \\
\midrule

\multirow{3}{*}{
{\begin{tabular}[c]{@{}c@{}} DINOv2 \\(Base) \cite{oquab2023dinov2}\end{tabular}}
} 
& Frozen                    &0.0        &64.8 &36.1 &57.4 &52.8 &\textbf{65.1} &40.2 &58.0 &54.4 \\
& Full-Tuning               &86.6       &62.2 &34.1 &54.3 &50.2 &58.5 &29.1 &60.6 &49.4\\
& \cellcolor{lightgreen}\textbf{CrossEarth-Gate}  &\cellcolor{lightgreen}1.4-1.8    & \cellcolor{lightgreen}\textbf{65.0} &\cellcolor{lightgreen}\textbf{38.0} &\cellcolor{lightgreen}\textbf{61.2} &\cellcolor{lightgreen}\textbf{54.7} &\cellcolor{lightgreen}64.5 &\cellcolor{lightgreen}\textbf{45.6} &\cellcolor{lightgreen}\textbf{62.3} &\cellcolor{lightgreen}\textbf{57.4} \\
\midrule

& Frozen & 0.0 &65.3 &37.0 &60.9 &54.4 &68.0 &43.0 &66.7 &59.2  \\
& Full-Tuning &304.2 &60.9 &29.9 &58.9 &49.9 &61.9 &32.9 &64.3 &53.0 \\
\rowcolor{lightgreen} \cellcolor{white}
& \textbf{CrossEarth-Gate} &3.0-4.4  &68.3 &46.3 &\textbf{63.3} &\textbf{59.3} &69.0 &50.0 &68.1 &\textbf{62.4}  \\
&w/o Spatial   &2.7-3.2 
&\textbf{70.0} &45.3 &60.5 &58.6 &68.8 &49.3 &67.9 
&61.9\\
&w/o Semantic  &4.2-4.5  &68.3 &43.1 &61.6 &57.6 &68.9 &44.7 &67.6 &60.4\\
&w/o Frequency &3.2-4.2  &67.7 &\textbf{47.4} &62.9 &\textbf{59.3} &\textbf{69.1} &45.5 &\textbf{68.3} &60.9\\
\multirow{-7}{*}{
{\begin{tabular}[c]{@{}c@{}} DINOv2\\(Large) \cite{oquab2023dinov2}\end{tabular}}} 
&w/o Selection &14.4  &69.5 &41.2 &61.2 &57.3 &68.3 &\textbf{51.0} &63.9 &61.1 \\

\bottomrule
\label{exp:completeablation}
\end{tabular}}
\end{table*}

%% file: sec/tab/tab6.tex
\begin{table*}[t]
    \centering
    \setlength\tabcolsep{3pt} 
    \caption{Detailed statistics and configuration of the cross-domain remote sensing benchmarks, including Domain Adaptation (DA) and Domain Generalization (DG). We categorize the experimental settings by the type of domain gap (Unseen Region, Spectral Band, and Climate) and provide specific details regarding the source and target domains, data splits (train/test numbers), image resolutions, and category counts for LoveDA \cite{wang2021loveda}, Potsdam, Vaihingen, RescueNet \cite{rahnemoonfar2023rescuenet}, and CASID \cite{liu2023large} datasets.}
    \resizebox{\textwidth}{!}{
    \begin{tabular}{cccccccccc}
    \toprule
    \multirow{1}{*}{Domain Gap} &Benchmark &\multicolumn{1}{c}{Dataset} & \multicolumn{1}{c}{Source Domain} &\multicolumn{1}{c}{Target Domain} & Abbreviation & Train Number & Test Number & \multicolumn{1}{c}{Image Size} & Categories \\
    \midrule
    \multirow{4}{*}{Unseen Region} & \multirow{2}{*}{DA}
    &\multirow{2}{*}{LoveDA\cite{wang2021loveda}}& LoveDA-Urban & LoveDA-Rural& U2R& 5464 & 3968 & \multirow{2}{*}{512$\thinspace\times\thinspace$512} &7 \\
    
    &&&LoveDA-Rural & LoveDA-Urban & R2U &5464 &2708 & & 7\\ & \multirow{2}{*}{DG} 
    &\multirow{1}{*}{Potsdam, RescueNet \cite{rahnemoonfar2023rescuenet}} & \multirow{2}{*}{Potsdam (RGB)} & \multirow{2}{*}{RescueNet (RGB)} & \multirow{2}{*}{P(r)$\thinspace$2$\thinspace$Res} & \multirow{2}{*}{3456} & \multirow{2}{*}{449} & \multirow{2}{*}{512$\thinspace\times\thinspace$512}&\multirow{2}{*}{5}\\
    && (Disaster Assessment)&&&&&&&\\
    \midrule
     \multirow{4}{*}{\begin{tabular}[c]{@{}c@{}} Unseen Region \\ and Spectral Band \end{tabular}} &\multirow{2}{*}{DA} &\multirow{2}{*}{Potsdam, Vaihingen}& Potsdam (RGB) & Vaihingen (IRRG) & P2V &3456$\thinspace$&398$\thinspace$ & \multirow{2}{*}{512$\thinspace\times\thinspace$512}&6\\
    && & Vaihingen (IRRG) & Potsdam (RGB) & V2P& 3456$\thinspace$& 2016$\thinspace$&  & 6 \\
    & \multirow{2}{*}{DG} & \multirow{1}{*}{Potsdam, RescueNet \cite{rahnemoonfar2023rescuenet}}& \multirow{2}{*}{Potsdam (IRRG)}  & \multirow{2}{*}{RescueNet (RGB)}& \multirow{2}{*}{P(i)$\thinspace$2$\thinspace$Res} & \multirow{2}{*}{3456} & \multirow{2}{*}{449} & \multirow{2}{*}{512$\thinspace\times\thinspace$512}&\multirow{2}{*}{5}\\
    & & \multirow{1}{*}{(Disaster Assessment)}  & & & & & & &
    \\
    \midrule
    \multirow{12}{*}{\begin{tabular}[c]{@{}c@{}} Unseen Region \\ and Climate \end{tabular}}& \multirow{12}{*}{DG} &\multirow{12}{*}{CASID \cite{liu2023large}} & \multirow{3}{*}{Subtropical Monsoon} & Temperate Monsoon& Sub2Tem& \multirow{3}{*}{4900}&2075 & \multirow{12}{*}{1024$\thinspace\times\thinspace$1024} & 5 \\
    & & & & Tropical Monsoon& Sub2Tms& &1650 & & 5\\
    & & & & Tropical Rainforest& Sub2Trf & &1550 & &5 \\
    & & &\multirow{3}{*}{Temperate Monsoon} &  Subtropical Monsoon&Tem2Sub&\multirow{3}{*}{5025} &2200 & & 5\\
    & & & & Tropical Monsoon& Tem2Tms& &1650 & &5 \\
    & & & & Tropical Rainforest& Tem2Trf & &1550 & &5 \\
    & & &\multirow{3}{*}{Tropical Monsoon} & Subtropical Monsoon& Tms2Sub&\multirow{3}{*}{3400}&2200 &  &5 \\
    & & & & Temperate Monsoon& Tms2Tem& &2075 & &5 \\
    & & & & Tropical Rainforest&Tms2Trf & &1550 & &5 \\
    & & &\multirow{3}{*}{Tropical Rainforest} & Subtropical Monsoon&Trf2Sub &\multirow{3}{*}{3700} &2200 &  &5 \\
    & & & & Temperate Monsoon& Trf2Tem& &2075 & &5 \\
    & & & & Tropical Monsoon& Trf2Tms & &1650 & &5 \\
    
    \bottomrule
    \end{tabular}}

    \label{benchmark}
\end{table*}

%% file: sec/tab/tab7.tex
\begin{table}[tbp]
\caption{Default hyperparameters of PEFT baselines.}
\begin{tabular}{lllc}
\toprule
Method & Hyperparameter & Meaning & \multicolumn{1}{c}{Value} \\ \midrule
\multirow{2}{*}{VPT \cite{jia2022visual}} & Prompt length & The number of soft prompt tokens & 128 \\
& Dropout & Dropout rate & 0\\ 
\midrule
\multirow{1}{*}{SLR \cite{scheibenreif2024parameter}} &Bottleneck dimension & The hidden dimension of the adapter & 16 \\
 \midrule
 \multirow{4}{*}{Rein \cite{wei2024stronger}} & Prompt length & The number of soft prompt tokens & 100 \\
& Rank & The rank of the low rank matrix & 16 \\
& Link to query & Whether to link the token to query &True \\
& Scale & The initial scale of the adapter feather & 0.001\\
 \midrule
 & Dropout & Dropout rate & 0.1\\
\midrule
\multirow{3}{*}{LoRA \cite{hu2022lora}} & Rank & The rank of the low rank matrix & 64 \\
 & Alpha & The alpha value & 1 \\
 & Dropout & Dropout rate & 0\\ 
 \midrule
  \multirow{2}{*}{Adapter \cite{houlsby2019parameter}} &Bottleneck dimension & The hidden dimension of the adapter & 64 \\  
 & Dropout & Dropout rate & 0.1\\
\midrule
  \multirow{3}{*}{AdaptFormer \cite{chen2022adaptformer}} &Bottleneck dimension & The hidden dimension of the adapter & 64 \\  
 & Dropout & Dropout rate & 0.1\\
 & Scale & The initial scale of the adapter feather & 0.1\\
 \midrule
 \multirow{10}{*}{Earth-Adapter \cite{hu2024learn}} &\multirow{3}{*}{Bottleneck dimension} & \multirow{3}{*}{The hidden dimension of the adapter} & 16 on R2U \\
 & & & 32 on U2R\\
 & & & 64 on other benchmarks\\
      & Scale & The initial scale of the adapter feather & 0.1\\
  & \multirow{2}{*}{Cut off ratio} & \multirow{2}{*}{The threshold to decompose frequncy}  & 0.2 on V2R \& V2P\\
  & & & 0.3 on other benchmarks\\
   & \multirow{4}{*}{Frequency layer} &\multirow{4}{*}{The layer to apply frequency Adapters} & [0-2] on P2V \\
   & & & [21-23] on V2P \& R2U\\
   & & & [18-23] on U2R\\
   & & & [0-23] on other benchmarks\\
  
\bottomrule
\end{tabular}
\centering
\label{baseline}
\end{table}

%% file: sec/tab/tab8.tex
\begin{table}[htbp]
\caption{Default hyperparameters of CrossEarth-Gate.}
\begin{tabular}{lllc}
\toprule
Method & Hyperparameter & Meaning & \multicolumn{1}{c}{Value} \\ \midrule
 \multirow{7}{*}{CrossEarth-Gate} & Active nodules & The number of selected modules & 18 \\
& Semantic dimension & The hidden dimension of the semantic module & 64 \\
& Spatial dimension & The hidden dimension of the spatial module & 64 \\
& Frequency dimension & The hidden dimension of the frequency module & 32\\
& Selection number & The number of the selections & 10 \\
& \multirow{2}{*}{Accumulation steps} & \multirow{2}{*}{Steps to accumulate Fisher information} & 50 on DA bechmarks \\
& & & 100 on DG bechmarks\\

\bottomrule
\end{tabular}
\centering
\label{gate}
\end{table}